\documentclass[journal]{IEEEtran}

\usepackage{ifpdf}
\usepackage{cite}
\ifCLASSINFOpdf
  \usepackage[pdftex]{graphicx}
\else
  \usepackage[dvips]{graphicx}
\fi
\usepackage{amsmath}
\usepackage{algorithmic}
\usepackage{array}
\ifCLASSOPTIONcompsoc
  \usepackage[caption=false,font=normalsize,labelfont=sf,textfont=sf]{subfig}
\else
  \usepackage[caption=false,font=footnotesize]{subfig}
\fi
\usepackage{fixltx2e}
\usepackage{stfloats}
\ifCLASSOPTIONcaptionsoff
  \usepackage[nomarkers]{endfloat}
 \let\MYoriglatexcaption\caption
 \renewcommand{\caption}[2][\relax]{\MYoriglatexcaption[#2]{#2}}
\fi
\usepackage{url}

\usepackage{times}
\usepackage{multicol}
\usepackage{amsbsy}
\usepackage{amsmath}
\usepackage{amssymb}
\usepackage{caption}
\usepackage{hhline}
\usepackage{multirow}
\usepackage{floatrow}
\usepackage{textcomp}
\usepackage{color}
\usepackage{xspace}
\usepackage{enumerate}
\usepackage{csquotes}
\usepackage{siunitx}
\usepackage{rotating}
\usepackage{booktabs}

\newlength{\Oldarrayrulewidth}
\newcommand{\Cline}[2]{
  \noalign{\global\setlength{\Oldarrayrulewidth}{\arrayrulewidth}}%
  \noalign{\global\setlength{\arrayrulewidth}{#1}}\cline{#2}%
  \noalign{\global\setlength{\arrayrulewidth}{\Oldarrayrulewidth}}}

\def\LatexSettings{./settings} 
\def\include{./include}

\input{\LatexSettings/def_macros.tex}

\begin{document}

\title{The Role of the Input in\\Natural Language Video Description}

\author{Silvia~Cascianelli,~\IEEEmembership{Student Member,~IEEE,}
		Gabriele~Costante,~\IEEEmembership{Member,~IEEE,} Alessandro~Devo,
		Thomas~A.~Ciarfuglia,~\IEEEmembership{Member,~IEEE,}
        Paolo~Valigi,~\IEEEmembership{Member,~IEEE,}
        and Mario~L.~Fravolini
\thanks{The authors are with the Department
of Engineering, University of Perugia, Perugia,
Italy (e-mail: silvia.cascianelli@unipg.it; gabriele.costante@unipg.it; alessandro.devo@studenti.unipg.it; thomas.ciarfuglia@unipg.it; paolo.valigi@unipg.it; mario.fravolini@unipg.it)}
\thanks{Manuscript received Month XX, 2018; revised Month XX, 201X.}}


\maketitle
\begin{abstract}
Natural Language Video Description (NLVD) has recently received strong interest in the Computer Vision, Natural Language Processing (NLP), Multimedia, and Autonomous Robotics communities. The State-of-the-Art (SotA) approaches obtained remarkable results when tested on the benchmark datasets. However, those approaches poorly generalize to new datasets. In addition, none of the existing works focus on the processing of the input to the NLVD systems, which is both visual and textual. In this work, it is presented an extensive study dealing with the role of the visual input, evaluated with respect to the overall NLP performance. This is achieved performing data augmentation of the visual component, applying common transformations to model camera distortions, noise, lighting, and camera positioning, that are typical in real-world operative scenarios. A t-SNE based analysis is proposed to evaluate the effects of the considered transformations on the overall visual data distribution. 
For this study, it is considered the English subset of Microsoft Research Video Description (MSVD) dataset, which is used commonly for NLVD. 
It was observed that this dataset contains a relevant amount of syntactic and semantic errors. These errors have been amended manually, and the new version of the dataset (called MSVD-v2) is used in the experimentation. The MSVD-v2 dataset is released to help to gain insight into the NLVD problem.
\end{abstract}
\begin{IEEEkeywords}
Video Description, Multimodal Data, Input Preprocessing.
\end{IEEEkeywords}

\section{Introduction}
\begin{figure}[]
\centering
\includegraphics[width =\columnwidth]{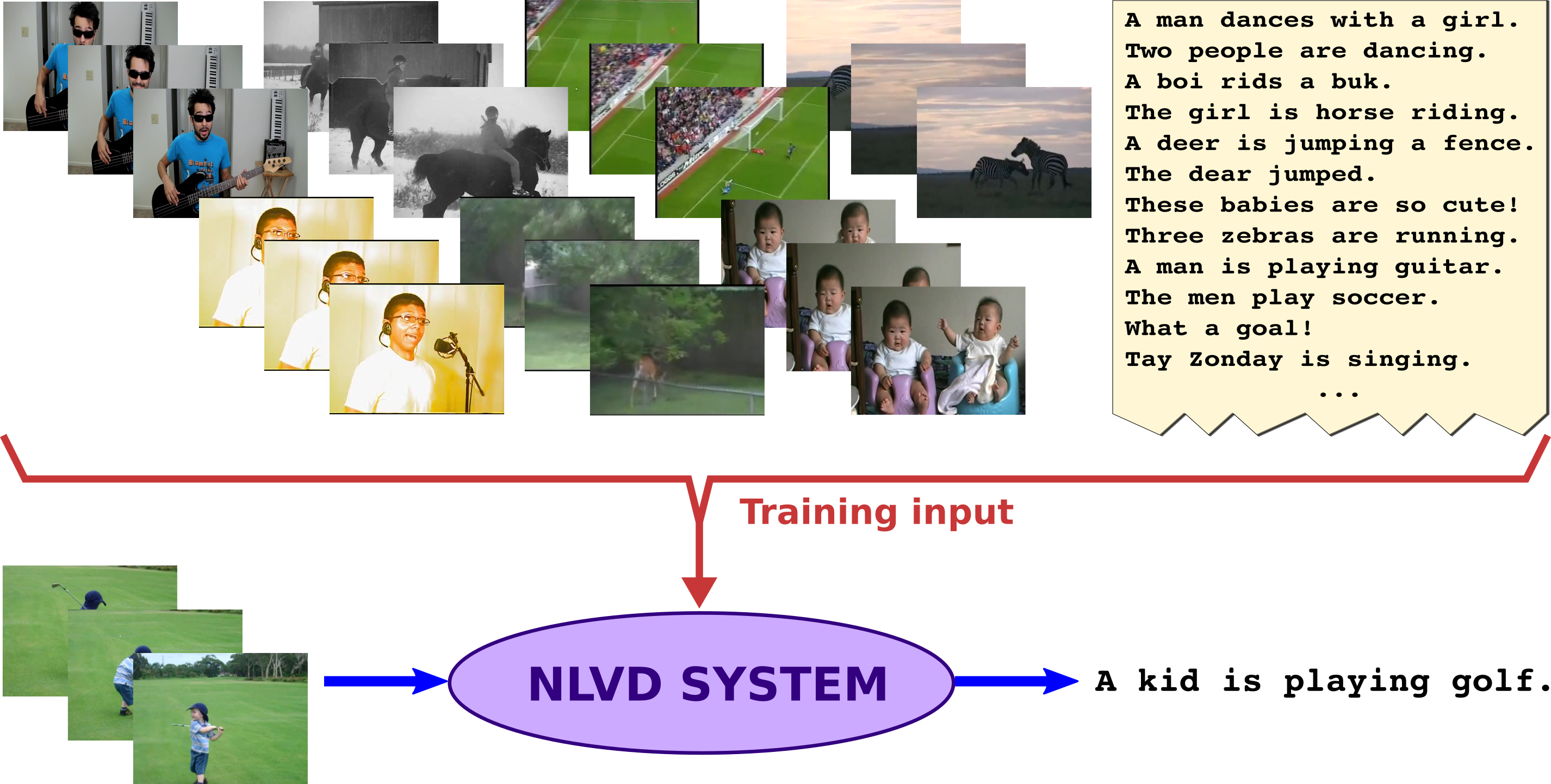}
\caption{Natural Language Video Description systems are trained on videos and associated captions. In the test phase, these systems are expected to produce a relevant and syntactically correct sentence describing unseen videos. }
\label{fig:overview}
\end{figure}

\IEEEPARstart{V}{isual and textual} data-based tasks \cite{cui2018general} are receiving growing interest in many research communities. Some studied problems are visual content retrieval based on natural language queries \cite{kofler2014predicting,xie2014contextual,li2017joint,hu2018twitter100k}, text-guided video summarization \cite{song2018extracting,yang2018text2video}, story understanding \cite{baraldi2017recognizing}, and visual content description \cite{li2018gla,gao2017video,dong2018predicting}.
This paper tackles the video description problem (NLVD). This is particularly interesting both for its research challenges and for its numerous possible applications. These include automatic video captioning of web content, automatic generation of the Descriptive Video Service (DVS) track of movies, products for the visually impaired and the blind, effective human-machine interaction, service and collaborative robotics applications, and video surveillance to name a few.
The approaches developed to address this problem are data-driven. In the training phase, the NLVD systems receive as input a video stream and an associated description, that is a sentence in natural language. In the test phase, those systems are expected to output a descriptive sentence given a video (\figname~\ref{fig:overview}). The quality of the produced description is difficult to assess objectively \cite{vedantam2015cider, anderson2016spice}. Nevertheless, to obtain a quantitative evaluation, the common practice is adopting metrics designed for NLP tasks such as machine translation and summarization, and for image description.
The SotA approaches obtained good results on the benchmark datasets in terms of evaluation metrics. However, the human performance in terms of the same metrics is still significantly higher (see \TABLEname~\ref{tab:human}). 
Another issue with the current NLVD methods is that both training and test are performed on the same dataset. The recent work by Cascianelli \etal \cite{cascianelli2018full} outlined the poor generalization capabilities of those algorithms when tested on a new dataset. This may limit their practical applicability. 

The recently growing interest in NLVD is accompanied by intense activity of design and collection of new datasets suitable for studying the problem. 
The most commonly used datasets for NLVD are the Montreal Video Annotation dataset (M-VAD) \cite{torabi2015using}, the Max Plank Institute of Informatics Movie Description dataset (MPII-MD) \cite{rohrbach2015dataset}, the Microsoft Video Description Corpus (MSVD) \cite{guadarrama2013youtube2text} and the Microsoft Research - Video to Text dataset (MSR-VTT) \cite{xu2016msr}. 
These datasets are generic in the depicted actions and featured actors in the scene. The M-VAD and the MPII-MD contain snippets from movies, which typically have high resolution. The MSVD and the MSR-VTT, instead, include videos from YouTube, which thus have a more varied quality.  In this respect, these two latter datasets seem more suitable for the study of NLVD systems able to generalize. However, it is not guaranteed that they capture the high variability of the video quality (\eg color channels, resolution) in the problem. This is an obstacle to the deploy of NLVD systems in applications such as surveillance and service robotics, where the characteristics of the camera and its position in the scene differ from scenario to scenario. 
From a textual standpoint, in the M-VAD and MPII-MD, the videos are paired with the associated sentence from the script or the transcribed DVS track. Therefore, these datasets lack in diversity of the possible description for each video. This also has a drawback in the evaluation procedure since using the standard evaluation metrics can be to some extent misleading \cite{rohrbach2017movie}.  In the MSVD and MSR-VTT, there are several descriptions for each video (on average 43 in the first dataset, 20 in the other,) collected via the Amazon Mechanical Turk (AMT) service.  Since they better capture the different ways to describe the same video, these two datasets seem more suitable to study NLVD. However, not only SotA methods still perform poorly on them, but also humans obtain not perfect performance scores (see \TABLEname~\ref{tab:human} and \TABLEname~\ref{tab:msrvtt}). In the sight of these considerations, this study is conducted using the MSVD dataset. 

It is well known that the quality of the training data is crucial for the performance of NLVD algorithms. Therefore, it is important to use the most reliable datasets, deeply analyse their characteristics, and design the training input properly. 
Input preprocessing is a well-known good practice for effectively training machine learning algorithms \cite{teng1999correcting, kotsiantis2006data, nawi2013effect}. For example, via data augmentation the training set can be automatically enlarged, thus providing more samples to the algorithm. This reduces the overfitting and increases the generalization capability of the model. Further, via data cleansing outliers and incorrect samples are removed, thus the distribution of the dataset should better represent the problem. This reduces the training time and increases the accuracy of the models. To the best of our knowledge, the role of the input has been neglected so far for NLVD systems. In our opinion, this aspect should be deeply explored for two main reasons: to allow improving the generalization capabilities and to gain further insights into the problem and thus design NLVD algorithms more judiciously.
In the sight of this, the purpose of this work is to tackle the following practical issues: 1) to quantify the performance improvement due to input preprocessing; 2) to provide some practical guidelines for a rational selection of suitable input augmentation strategies.

For this study, the benchmark MSVD dataset is considered, and a standard encoder-decoder NLVD system is designed. A number of visual transformations are then applied to the videos in the dataset. The selection of the most appropriate appearance transformations for visual data augmentation is guided both by a data-driven analysis based on t-SNE [23]. Further, since the transformed videos have to preserve the original semantic content, the augmentation strategies have been selected among those that do not affect the relation between the video and the associated description. In the experimentation, it was observed that the MSVD dataset contains a relevant number of syntactic and semantic errors. This suggested to (manually) amend these inconsistencies producing an improved dataset, called MSVD-v2. This new dataset is used in addition to the original one in the experiments, to evaluate the effects of training the NLVD system with more consistent textual data. 

The remainder of the paper is organized as follows. In Section \ref{sec:related_work} the related work is overviewed. In Section \ref{sec:approach} the poposed approach is explained. In Section \ref{sec:experiments} the result of an extensive experimental study are reported and discussed. In Section \ref{sec:conclusion} the conclusions are traced.

\section{Related Work}\label{sec:related_work}
The NLVD problem is attracting the interest of many research communities, from the Computer Vision \cite{torabi2015using} and NLP \cite{thomason2014integrating} community to the Multimedia \cite{cho2015describing,gao2017video,dong2018predicting} and Autonomous Robotics ones \cite{cascianelli2018full}.
The early proposed approaches to NLVD consist in addressing the task as template filling \cite{krishnamoorthy2013generating, guadarrama2013youtube2text, thomason2014integrating} or description retrieval \cite{zhu2015aligning, dong2018predicting}.

The most recent and most popular approach to NLVD is treating the problem as a machine translation one \cite{venugopalan2015sequence}, from a video sequence to a natural language sentence, using and encoder-decoder architecture. 
The frames of the video are usually subsampled and processed by one or more Convolutional Neural Network (ConvNet) to extract a visual descriptor for the frame. Object recognition  ConvNets and action recognition ConvNets are commonly used and combined together to obtain a good representation of the frames. Integrating the Optical Flow is also a used strategy \cite{guo2018exploiting}. Another recently proposed approach \cite{xu2018sequential} consists in representing the video frames via a sequential vector of locally aggregated descriptor (SeqVLAD) layer, that combines a VLAD encoding and a recurrent-convolutional network. The SeqVLAD framework aggregates the intra-frame spatial information and the inter-frame motion information. 
The frames descriptors are used to encode the video. The econding can be obtained directly by mean pooling the features, as done, \eg in \cite{venugopalan2014translating}, or, more effectively, via an RNN-based encoder. Typically, is used an LSTM-based encoder. This can be a single LSTM \cite{li2018residual}, a bidirectional LSTM (BiLSTM) \cite{bin2018describing}, or a multilayer LSTM \cite{baraldi2017hierarchical}. 
Using the GRU in the encoder is less common \cite{cascianelli2018full}.
The video encoding is then fed to the sentence decoder together with the ground truth sentence, word-by-word. The words in the ground truth sentences are used to form a vocabulary for the dataset.
The words in the caption are represented as vectors in a Word Embedding (WE) \cite{mikolov2013efficient, pennington2014glove}. The WE is usually learned during the training of the NLVD system \cite{song2017deterministic}, or in some cases is a pretrained WE, as in \cite{venugopalan2016improving}. 
The decoder is trained to predict the probability of each word in the vocabulary to be the next one in the sentence based on the video encoding and the previous words in the sentence. At each step, the most probable word is emitted, and the process stops when an End-Of-Sequence ($<$EOS$>$) tag is emitted. The decoder is designed to be a recurrent architecture. The LSTM is the preferred choice, either as a single block \cite{rohrbach2017movie} or in a multilayer LSTM-based architecture \cite{guo2018exploiting}.
Some works \cite{baraldi2017hierarchical, cascianelli2018full, wang2018sequence} employ the GRU as the main block of the decoder. 

To improve the performance, attention mechanisms are employed at different points of the encoder-decoder system. 
In particular, at each word generation step, the decoder takes as input the video features weighted according to their relevance to the next word, based on the previously emitted words \cite{bin2018describing, li2018multimodal, li2018residual, wang2018sequence}. 
With the same principle, in \cite{chen2016video} the attention mechanism is applied to the mean-pooled features from a predefined number of objects tracklets in the video. In \cite{yang2017catching}, the textual information is used to select Regions-of-Interest (ROIs) in the video frames, whose descriptors are combined with those of the global frame in a Dual Memory Recurrent Model. An alternative strategy to combine visual and textual information is reshaping the feature vectors into circulant matrices and combining them to extract the multimodal relation among the two different modalities \cite{wu2018multi}, or builnding multimodal matching tensor of sequential data \cite{yu2018joint}. 
The attention mechanism can be implemented as an additional layer in the encoder-decoder architecture or can be integrated into the gating strategy of the decoder, as done in \cite{gao2017video}.
Recent trends include training multitask NLVD models \cite{pasunuru2017multi, li2018end}, using a reinforcement-learning framework \cite{pasunuru2017reinforced, wang2018video}, or a cycle learning framework \cite{wang2018reconstruction}.

Devising a SotA NLVD system is beyond the scope of this paper. Here, the focus is on the input to these systems and the effects of its preprocessing on the NLVD performance. The study is conducted considering a simple yet effective NLVD encoder-decoder architecture.
\paragraph{Input Preprocessing}
Data-driven approaches, such as Deep Learning-based ones, heavily depend on the quality of the training data, in terms of effectiveness, achieved representation power, and generalization capability. For this reason, attention is usually put on properly preprocessing the input to those algorithms \cite{kotsiantis2006data}. 
Data augmentation at the visual level is a well-known strategy to improve the performance of algorithms for many Computer Vision tasks. 
Emblematic is the case of \cite{krizhevsky2012imagenet}, where the generalization capabilities of the AlexNet ConvNet increased by training the model on altered images. 
To be beneficial for the training, the applied alterations should be carefully designed to capture the characteristics of the data of the problem. In this work, it is proposed for the first time visual data augmentation for NLVD, taking into account the characteristics of the videos captured by the camera in various application scenarios, and maintaining the relation with the associated descriptions. 

In the recent work in \cite{jackson2018style}, it is presented style augmentation as a novel strategy to perform visual data augmentation exploiting a style transfer network \cite{ghiasi2017exploring}. In particular, the texture, contrast, colour and illumination of the image is altered, but shapes and semantic content are preserved. This strategy has been found effective for improving the performance on classification tasks, domain transfer and depth estimation. Style transfer via neural networks was introduced by Gatys \etal in \cite{gatys2016image}, and many other works followed this approach for transforming images with the style of paintings \cite{johnson2016perceptual, vedaldi2016instance, ghiasi2017exploring} or other photorealistic images taken under completely different conditions \cite{li2018closed}. The content representation and the style representation of the input image are extracted from a pre-trained ConvNet. In particular, the content is represented by the feature responses in higher layers, and the style is represented by the feature correlations of multiple lower layers. Content and style are modelled by two separate terms of the loss function, minimized to synthesise the new image having the desired style and content.
Following the novel approach of \cite{jackson2018style}, in this work style augmentation is tested for NLVD.

In the NLP literature, data augmentation has been proposed to enlarge the training corpora automatically. For example, the authors of \cite{zhang2015text} performed textual data augmentation by replacing words with their synonyms from WordNet \cite{fellbaum2010wordnet} for ConvNet-based models for ontology classification, sentiment analysis, and text categorization. In \cite{saito2017improving} the focus was on Natural Language Normalization and it was addressed the problem of small datasets for that. The authors trained a machine translation architecture on a small normalization dataset and translated in an unnormalized form a bigger corpus of standard text. With this, the authors were able to augment the small text normalization datasets. In \cite{fadaee2017data}, data augmentation for machine translation was performed, targeting rare words. The authors trained an LSTM language model to alter both source and target sentences in a parallel corpus. This way, they maintained the relation between the two sentences in the two languages. Doing the same for NLVD is not straightforward because one of the two "languages" is visual. 
Few works on NLVD operate at input level. In \cite{chen2016video} data augmentation is proposed at the sentence level. The authors proposed to enrich the sentence part of the MSR-VTT with sentences from the MSVD. These sentences are selected based on the visual similarity between the associated videos in the two datasets. However, once included in the MSR-VTT, the sentences are paired with fake videos, \ie all-zeros vectors. Thus, this approach does not maintain the relation between video and text. In this paper, a new version of a benchmark dataset is presented. The sentences associated with the videos have been manually checked and corrected in case of errors, thus maintaining their semantic relatedness to the videos.
\section{Proposed Approach}\label{sec:approach}
To study the role of the input in the NLVD problem a basic encoder-decoder architecture is designed, and a standard benchmark dataset, namely the MSVD \cite{guadarrama2013youtube2text},  isn considered. In this section, it is described the NLVD system, the video augmentation strategy, and the text checking procedure that led to the amended version of the dataset.

First of all, it is instructive to briefly overview the standard evaluation metrics used for NLVD systems and throughout this study to guide the design choices. These metrics are: BLEU \cite{papineni2002bleu}, in its \emph{4}-gram variant; ROUGE \cite{lin2004rouge} in its Longest Common Subsequence (LCS) variant; METEOR \cite{banerjee2005meteor}; CIDEr \cite{vedantam2015cider}.

Call \emph{n}-gram a sequence of \emph{n} consecutive words. Given a candidate sentence A and a reference sentence B to compare:
\begin{itemize}
\item The ratio of the number of \emph{n}-grams in A that are mapped to \emph{n}-grams in B to the total number of \emph{n}-grams in A is the \emph{n}-gram precision.
\item The ratio of the number of \emph{n}-grams in A that are mapped to \emph{n}-grams in B to the total number of \emph{n}-grams in B is the \emph{n}-gram recall.
\end{itemize}

BLEU is a precision-oriented metric designed for machine translation evaluation. To obtain the score, \emph{n}-gram precision is calculated considering \emph{n}-grams up to length four.
BLEU correlates well with human judgement on the quality of the translation when evaluated on the entire test set, but poorly at the sentence level. 

ROUGE is a recall-oriented metric designed for summarization evaluation. It is based on the idea that a candidate summary should ideally overlap the reference summary. This metric has three variants, depending on the sentences comparison strategy adopted. In the NLVD literature, it is used the variant that considers the longest common sequence (LCS), called ROUGE$_L$.
All ROUGE variants correlate well with human judgement.

METEOR is a precision and recall-based MT evaluation metric. For its computation, unigrams in the candidate and reference sentences are matched based on their exact form, \ie if the unigrams are the same word, stemmed form, \ie if the unigrams have the same root, and meaning, \ie, if the unigrams are synonyms. Then, unigram precision and unigram recall are calculated based on the found matches, and the F-mean is obtained, weighing the recall more than the precision. In addition, a multiplicative factor is used to reward identically ordered contiguous matched unigrams. METEOR correlates better than unigram precision, unigram recall and their harmonic combination, with human judgement also at the sentence level.

CIDEr is a metric designed to assess the quality of the description of an image. It is based on the cosine similarity between \emph{n}-grams in the candidate description and in the set of reference descriptions associated to the image. Each \emph{n}-gram is weighted using a Term Frequency-Inverse Document Frequency (TF-IDF) strategy.
This metric is designed to correlate well with human judgement on the image description quality, thus is particularly suitable for the task of NLVD. 

The possible values for all the above metrics span from 0 to 1. For all but CIDEr, these are reported using values from 0 to 100. The values of the CIDEr metric are reported between 0 and 1000. This is done to make the CIDEr values of the same order of magnitude as those of the other metrics. In fact, even SotA approaches obtain very low scores in terms of the CIDEr metric. 

\subsection{Basic Encoder-Decoder NLVD System}\label{ssec:bedds}
\begin{figure*}[t]
\centering
\includegraphics[width = \textwidth]{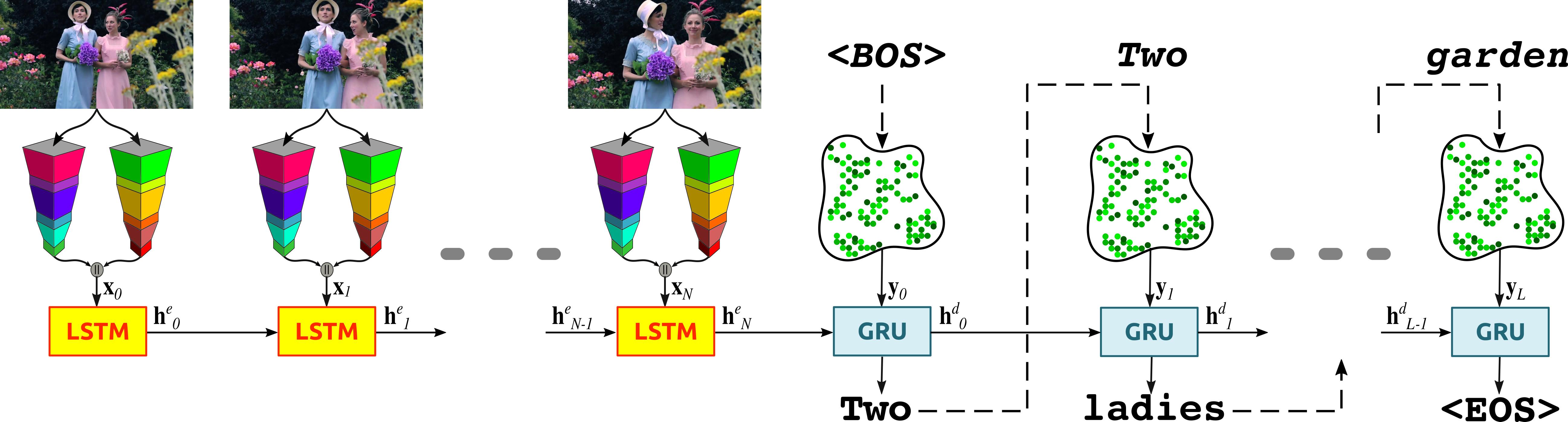}
\caption{Architecture of the encoder-decoder model used in this study. Recurrent layers are depicted as unfolded graphs for explanatory purpose. The \emph{ResNet50} and \emph{C3D} ConvNets extracts features from the video frames, which are the input to the LSTM encoder. The final state of the encoder and the GloVe embedding of the words in the caption are the input to the GRU decoder, which generates the output description one word at a time until it emits the $<$EOS$>$ tag. \vspace{-1.25em}}
\label{fig:model}
\end{figure*}

Outperforming the SotA is beyond the scope of this paper, thus a simple yet effective encoder-decoder architecture is designed and used. This helps in better highlighting the effects of the input preprocessing on the performance. Its pictorial representation is in \figname~\ref{fig:model}. In the following, the model is referred to as Basic Encoder-Decoder Description System (BEDDS).

\begin{table}[]
\centering
\caption{Preliminary ablation study on the encoder-decoder architecture used for this study on the MSVD. $B_4$ stands for BLEU\textsubscript{4}, $R_L$ for ROUGE\textsubscript{L}, $M$ for METEOR, and $C$ for CIDEr. Bold indicates the best performance. \vspace{-1.25em}}
\label{tab:preliminary}
\resizebox{\textwidth}{!}{%
\begin{tabular}{l|c|c|c|c|}
\cline{2-5}
 & $B_4$ & $R_L$ & $M$ & $C$ \\ \hline
\multicolumn{1}{|l|}{BEDDS (VGG16) + WE + VE} & 41.5 & 66.8 & 30.4 & 60.7 \\ \hline
\multicolumn{1}{|l|}{BEDDS (VGG16) + WE} & 41.2 & 67.0 & 30.9 & 57.1 \\ \hline
\multicolumn{1}{|l|}{BEDDS (VGG16) + WE - GRU enc.} & 41.9 & 67.5 & 30.6 & 54.1 \\ \hline
\multicolumn{1}{|l|}{BEDDS (ResNet50+C3D) + WE} & 45.0 & 69.2 & 32.3 & 66.7 \\ \hline
\multicolumn{1}{|l|}{BEDDS (ResNet50+C3D) + WE - GRU enc.} & 43.9 & 69.1 & \textbf{32.9} & 69.9 \\ \hline
\multicolumn{1}{|l|}{BEDDS (ResNet50+C3D) + GloVe finetuned} & 43.6 & 69.0 & 32.3 & 69.9 \\ \hline
\multicolumn{1}{|l|}{BEDDS (used for the study)} & \textbf{45.1} & \textbf{69.4} & \textbf{32.9} & \textbf{70.0} \\ \hline
\end{tabular}%
}
\end{table}

The video frames are sampled one every five. On the sampled frames, the output of the last fully connected layer of the \emph{ResNet50} \cite{he2016deep} and the \emph{C3D} \cite{tran2015learning} ConvNets is computed. The choice of these two SotA ConvNets is the result of a preliminary ablation study and confirms the results reported, \eg in \cite{guo2018exploiting, wang2018reconstruction} on the benefits of using very deep object recognition ConvNets and including the temporal information either via action recognition ConvNets or Optical Flow. This allows capturing both the appearance and the movement in the frame. In this study, it has been used \emph{ResNet50} instead of its deeper variants to limit the computational cost of the experiments. Note that for the \emph{C3D} vector it is considered a sliding window centered in the sampled frame containing 16 frames. The outputs of the ConvNets are concatenated to form the feature vector $\textbf{x}_\ast$ describing the frame. This vector is \num{2048}+\num{4096}-dimensional. As a result, the input video is represented by the sequence of feature vectors describing its $N$ frames $(\textbf{x}_0, \textbf{x}_1, ..., \textbf{x}_N)$. Usually, in the NLVD literature, the visual feature vectors are mapped in a lower dimensional space via a learnt linear transformation (VE). In the NLVD architecture used for this study, it has been decided not to perform this mapping operation in the sight of the preliminary study whose results are reported in \TABLEname~\ref{tab:preliminary}.

The sentence words are converted to lower-case, and the punctuation is removed. The Begin-Of-Sequence ($<$BOS$>$) and the $<$EOS$>$ tags are prepended and appended respectively to the sentence. Afterwards, the so preprocessed sentence is tokenized, and the tokens form the dataset vocabulary $D$. Some SotA NLVD approaches include in the vocabulary only those words that appear in the dataset with a minimum frequency. For this study, it is decided to include all the words in the vocabulary to exclude the effects of the additional minimum frequency hyperparameter on the performance. 
The words in the dataset are represented using the \num{300}-dimensional GloVe \cite{mikolov2013efficient, pennington2014glove} WE, pre-trained on a six billion words corpus. 
In many SotA architectures the WE is learned from scratch or a pretrained WE is finetuned during the training of the NLVD system. In this study, all these strategies for the WE have been tested, and the pretrained GloVe WE led to the best performance (see \TABLEname~\ref{tab:preliminary}.) In addition, with this choice, the overall model has fewer parameters to train. Note that in case a word in the dataset has not a corresponding vector in the GloVe embedding, a 300-dimensional random valued vector is assigned to it. In general, such words are either proper nouns, typos or very rare words. In fact, their amount decreases from $\sim$\num{2600} to $\sim$\num{130} after the textual data cleansing procedure described in \ref{ssec:text_cleaning}. 
As a result, the input sentence is represented by the sequence of embedding vectors corresponding to its $L$ words $(\textbf{y}_0, \textbf{y}_1, ..., \textbf{y}_L)$.

The frames feature vectors are fed, one at a time, to the encoder LSTM \cite{hochreiter1997long}. Using the LSTM as the main block of the encoder in the NLVD systems is a common practice. In the case of this study, the choice was guided by a preliminary study in which the LSTM and the GRU have been compared as main block of the encoder. The study (see \TABLEname~\ref{tab:preliminary}) confirmed the results of \cite{cascianelli2018full} in that the two blocks are equivalent in terms of overall performance. Although GRU has fewer parameters, for this work it is decided to use an LSTM-based encoder, because this is the common strategy in the NLVD literature.
The LSTM is a Deep-RNN able to handle both long and short-term temporal dependencies between serial data. It has an inner memory cell $\textbf{c}_n$ and a gating strategy to update the memory cell value and produce the output $\textbf{h}^e_n$, based on the current input $\textbf{x}_n$, and the previous state $\textbf{c}_{n-1}$ and output $\textbf{h}^e_{n-1}$. In particular, the new memory cell value is obtained by combining the previous value, multiplied by the forget gate $\textbf{f}_n$, and a candidate new state $\tilde{\textbf{c}}_n$, multiplied by the input gate $\textbf{i}_n$. This is to modulate how much to forget the previous value, and how much to update the current value with the new information. The output is obtained by multiplying the current memory cell with the output gate, that modulates how much memory to expose for the output. 
More formally, the LSTM used as the encoder in this study is defined by the following equations (\ref{eq:3})-(\ref{eq:8}).
\begin{equation}\label{eq:3}
\textbf{f}_n=\sigma(W_f\textbf{x}_n + U_f\textbf{h}^{e}_{n-1}+\textbf{b}_f)
\end{equation}
\begin{equation}\label{eq:4}
\textbf{i}_n=\sigma(W_i\textbf{x}_n + U_i\textbf{h}^e_{n-1}+\textbf{b}_i)
\end{equation}
\begin{equation}\label{eq:5}
\textbf{o}_n=\sigma(W_o\textbf{x}_n + U_o\textbf{h}^e_{n-1}+\textbf{b}_o)
\end{equation}
\begin{equation}\label{eq:6}
\tilde{\textbf{c}}_n=tanh(W_c\textbf{x}_n + U_c\textbf{h}^e_{n-1}+\textbf{b}_c)
\end{equation}
\begin{equation}\label{eq:7}
\textbf{c}_n=\textbf{f}_n\odot\textbf{c}_{n-1}+i_n\odot\tilde{\textbf{c}}_n
\end{equation}
\begin{equation}\label{eq:8}
\textbf{h}^e_n=\textbf{o}_n\odot tanh(c_n)
\end{equation}
where the $W_{\ast}$s, the $U_{\ast}$s, and $\textbf{b}_\ast$s are learnable weight matrices and bias vectors, $\sigma$ is the sigmoid function, $tanh$ is the hyperbolic tangent function, and $\odot$ is the element-wise product. 

The last output of the encoder, which represents the entire video, is passed to the decoder as its first state, \ie $\textbf{h}^e_N = \textbf{h}^d_0 \doteq \textbf{v}$. The first input to the decoder is the WE of the $<$BOS$>$ token, the subsequent inputs are the WE of the words in the sentence, which terminates with the WE of the $<$EOS$>$ token. In this work, the decoder is a GRU \cite{chung2014empirical}.
The GRU is a more recent Deep-RNN, able to deal with both long and short-term time dependencies between the elements in a sequence. Different from the LSTM, it has not a memory cell, and its output corresponds to its inner state $\textbf{h}^d_l$. Similar to the LSTM, the state is calculated via a gating strategy using the current input $\textbf{y}_l$ and the previous state $\textbf{h}^d_{l-1}$. First, a candidate state $\tilde{\textbf{h}}^d_l$ is computed from the current input and the previous state, multiplied by the reset gate $\textbf{r}_{l}$. This gate controls how much of the old state to forget in the candidate new state. Afterwards, the state is updated, also obtaining the output. To this end, the previous state and current candidate state are combined after being multiplied by the update gate $\textbf{z}_{l}$. This gate controls how much of the old state to maintain and how much of the current candidate state to use in the new state.
More formally, the GRU used as the decoder in this study is defined by the following equations (\ref{eq:10})-(\ref{eq:13}).
\begin{equation}\label{eq:10}
\textbf{r}_l=\sigma(W_r\textbf{y}_l+U_r\textbf{h}^d_{l-1}+\textbf{b}_r)
\end{equation}
\begin{equation}\label{eq:11}
\textbf{z}_l=\sigma(W_z\textbf{y}_l+U_z\textbf{h}^d_{l-1}+\textbf{b}_z)
\end{equation}
\begin{equation}\label{eq:12}
\tilde{\textbf{h}}^d_l = tanh(W_h\textbf{y}_l+U_h(\textbf{r}_l\odot\textbf{h}^d_{l-1})+\textbf{b}_h)
\end{equation}
\begin{equation}\label{eq:13}
\textbf{h}^d_l=(1-\textbf{z}_l)\odot\textbf{h}^d_{l-1}+z_l\odot\tilde{\textbf{h}}^d_l
\end{equation}
where the $W_{\ast}$s, the $U_{\ast}$s, and $\textbf{b}_\ast$s are learnable weight matrices and bias vectors.

At each step, the decoder outputs the state $\textbf{h}^d_l$. This is multiplied by a weight matrix $W_D$ to obtain the output vector $\hat{\textbf{y}}_l$. From this, the output word is selected from the vocabulary using the softmax function, that models the probability that the output word is the next one in the description, \ie:
\begin{equation}\label{eq:30}
Pr_(\hat{\textbf{y}}_l\mid\textbf{v},\textbf{y}_0,\textbf{y}_1,...,\textbf{y}_{l-1})\sim\frac{e^{\hat{\textbf{y}}_l}}{\sum\limits_{\textbf{y}\in D}e^{y}}
\end{equation}
In the training phase, the objective is to maximize the log-likelihood of the words over the sentence, \ie 
\begin{equation}\label{31}
\max\limits_{\textbf{W}} \sum\limits_{l=1}^{L}logPr_(\hat{\textbf{y}}_l\mid\textbf{v},\textbf{y}_0,\textbf{y}_1,...,\textbf{y}_{l-1})
\end{equation}
where $\textbf{W}$ denotes all the parameters of the model. 

In the test phase, the input to the GRU decoder at each step is the previous word emitted, and the decoding process stops automatically when the $<$EOS$>$ token is emitted as the most probable token.

\subsection{Data Augmentation to Study the Role of the Visual Input}\label{ssec:visual_DA}
For the training of Deep Neural Networks, the availability of a large number of training samples is critical. NLVD systems are no exception, as it can be read from \TABLEname~\ref{tab:n_videos}. This is a first reason to perform data augmentation for enlarging the training set via videos alteration. 
In addition, these systems lack in the generalization capability, as observed from \cite{cascianelli2018full}. 

In this work, to study the generalization capabilities and the performance of the designed NLVD system under not ideal characteristics of the visual input, it is proposed to apply alterations to the videos in the MSVD that could reflect some operating conditions of cameras in real-world scenarios. In fact, when the video captioning system is used in a specific application context (\eg for a greyscale surveillance camera placed above the monitored scene) some considerations on the characteristics of the images can be traced (the images will be greyscaled, keystone distorted, possibly blurred, occasionally very dark or very bright, etc.) According to those considerations, altered videos can be included in the dataset used for training or fine-tuning the captioning system.

\begin{figure*}[]
  \ffigbox{}
  {
  \CommonHeightRow
    {
      \begin{subfloatrow}[4]
    \ffigbox[\FBwidth]
    {\includegraphics[height=\CommonHeight]{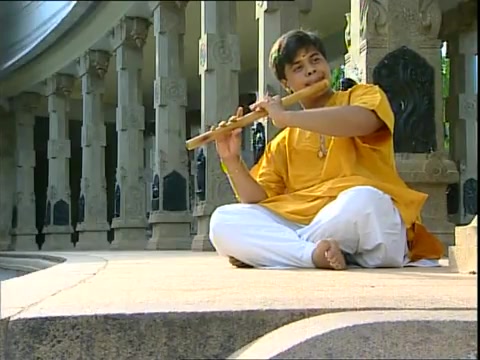}}
    {\caption{}\label{fig:original}}
    \ffigbox[\FBwidth]
    {\includegraphics[height=\CommonHeight]{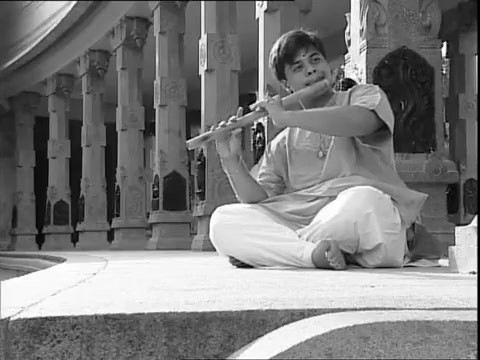}}
    {\caption{}\label{fig:gray}}
    \ffigbox[\FBwidth]
    {\includegraphics[height=\CommonHeight]{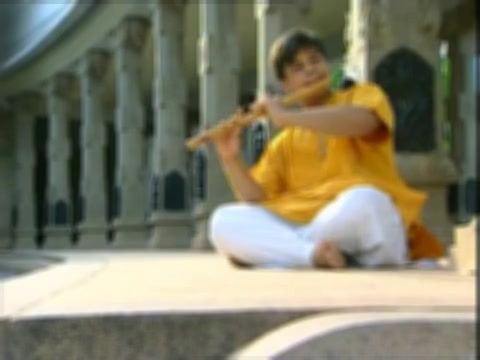}}
    {\caption{}\label{fig:blur}}
    \ffigbox[\FBwidth]
    {\includegraphics[height=\CommonHeight]{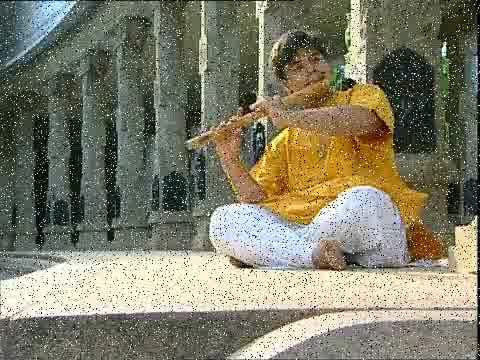}}
    {\caption{}\label{fig:sep}}
      \end{subfloatrow}
      \begin{subfloatrow}[4]
    \ffigbox[\FBwidth]
    {\includegraphics[height=\CommonHeight]{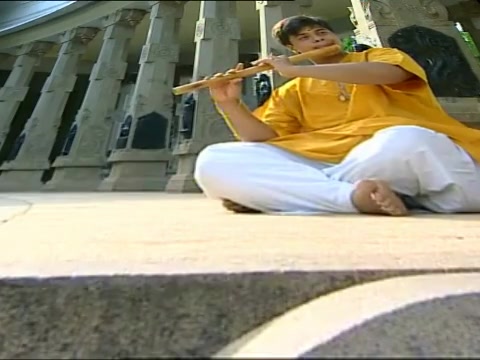}}
    {\caption{}\label{fig:td}}
    \ffigbox[\FBwidth]
    {\includegraphics[height=\CommonHeight]{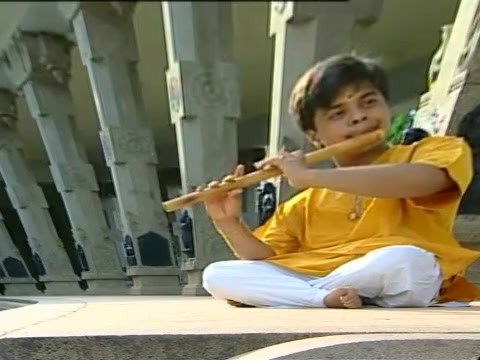}}
    {\caption{}\label{fig:bu}}
    \ffigbox[\FBwidth]
    {\includegraphics[height=\CommonHeight]{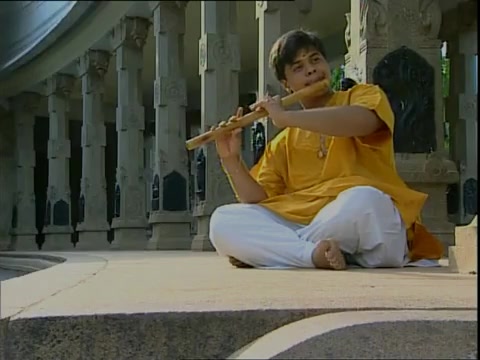}}
    {\caption{}\label{fig:dark}}
    \ffigbox[\FBwidth]
    {\includegraphics[height=\CommonHeight]{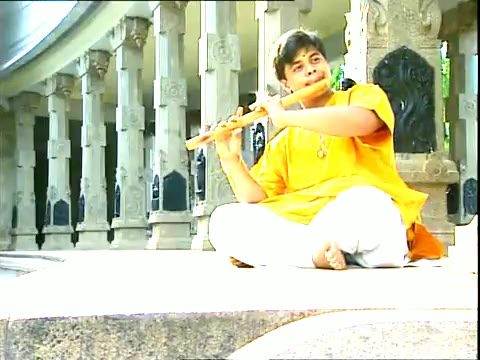}}
    {\caption{}\label{fig:light}}
      \end{subfloatrow}
    } 
\caption{Example of visual alterations applied to the training videos in the MSVD. \ref{fig:original} is an original image from the video. \ref{fig:gray} is the grayscale converted image. \ref{fig:blur} is the blurred image. \ref{fig:sep} is the image with applied gaussian noise (salt and pepper noise). \ref{fig:td} and \ref{fig:bu} are the image with two keystone distortions applied. \ref{fig:dark} is the brightness reduced image. \ref{fig:light} is the brightness enhanced image.}\label{fig:alterations}
	}
\end{figure*}

The transformations applied in this study are the following:
\begin{itemize}
\item Grayscale conversion, to model grayscale cameras, which are largely used \eg for surveillance and robotics applications.
\item Gaussian blur, to model the occasional out-of-focus operating condition.
\item Keystone distortion, to model the not optimal position of the camera in the scene. In fact, \eg flying drones and surveillance cameras are usually above the scene, while \eg small terrestrial robots, kids, or users seated underneath a stage are below the scene. In this work, this distortion has been implemented using the perspective transform.
\item Brightness enhancement and reduction, to model the different illumination conditions that may be encountered in the application scenario.
\item Salt \& Pepper noise, to generally represent low-quality images from cheap cameras.
\end{itemize}
Each of them is applied, to different degrees of severity, to all the videos on the MSVD. Exemplar applied transformations are reported in \figname~\ref{fig:alterations}.

\begin{table}[]
\centering
\caption{BEDDS model performance on the MSVD depending on the number of training videos. $B_4$ stands for BLEU\textsubscript{4}, $R_L$ for ROUGE\textsubscript{L}, $M$ for METEOR, and $C$ for CIDEr. Bold indicates the best performance. \vspace{-1.25em}}
\label{tab:n_videos}
\resizebox{\textwidth}{!}{%
\begin{tabular}{|c|c|cccc|}
\hline 
Training videos & Training samples & $B_4$ & $R_L$ & $M$ & $C$ \\ \hline
200 & 8182 & 33.0 & 63.7 & 27.6 & 36.3 \\ \hline
400 & 16221 & 35.4 & 65.6 & 29.3 & 48.7 \\ \hline
600 & 24599 & 41.6 & 66.3 & 29.6 & 52.4 \\ \hline
800 & 32606 & 42.0 & 68.0 & 31.1 & 58.4 \\ \hline
1000 & 41035 & 44.8 & 69.2 & 32.5 & 68.1 \\ \hline
1200 & 49158 & \textbf{45.1} & \textbf{69.4} & \textbf{32.9} & \textbf{70.0} \\ \hline
\end{tabular}%
}
\end{table}

\begin{figure*}[]
\centering
\ffigbox[]{%
\begin{subfloatrow}
  \ffigbox[\FBwidth][]
    {\caption{}\label{fig:or}}
    {\includegraphics[width=0.3915\textwidth]{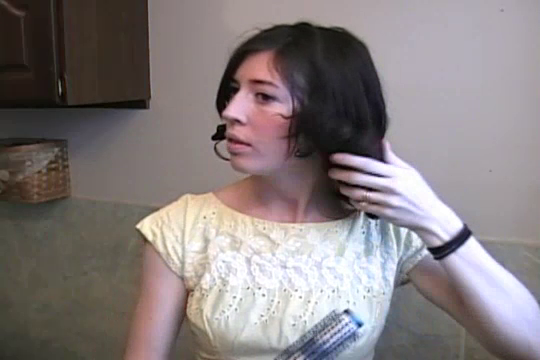}}
\end{subfloatrow}
\hspace*{0.01\textwidth}
\begin{subfloatrow}
	\vbox{
	\hbox{
    \ffigbox[\FBwidth]
    {\includegraphics[width=0.17\textwidth]{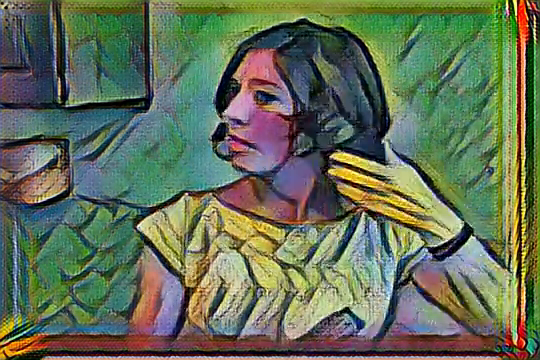}}
    {\caption{}\label{fig:muse}}
    \ffigbox[\FBwidth]
    {\includegraphics[width=0.17\textwidth]{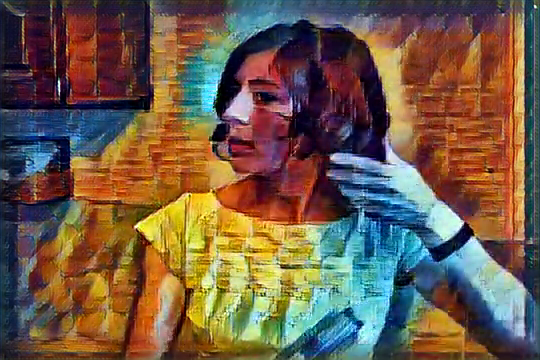}}
    {\caption{}\label{fig:rain}}
    \hspace*{0.015\textwidth}
    \ffigbox[\FBwidth]
    {\includegraphics[width=0.17\textwidth]{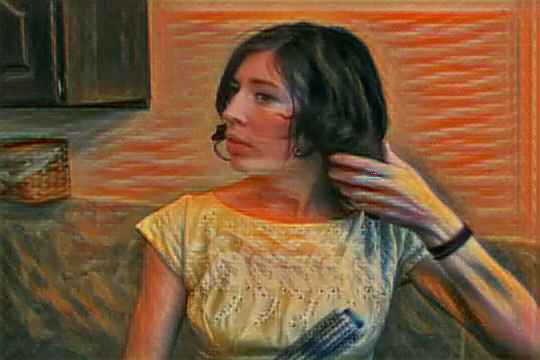}}
    {\caption{}\label{fig:scream}}
    }\vss
    \hbox{
    \ffigbox[\FBwidth]
    {\includegraphics[width=0.17\textwidth]{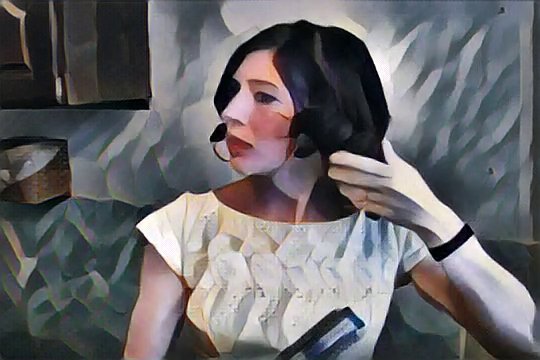}}
    {\caption{}\label{fig:udnie}}
    \ffigbox[\FBwidth]
    {\includegraphics[width=0.17\textwidth]{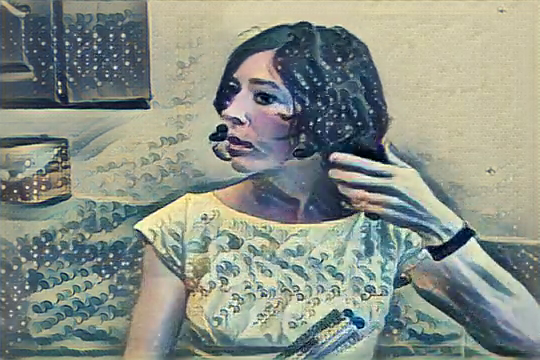}}
    {\caption{}\label{fig:wave}}
    \ffigbox[\FBwidth]
    {\includegraphics[width=0.17\textwidth]{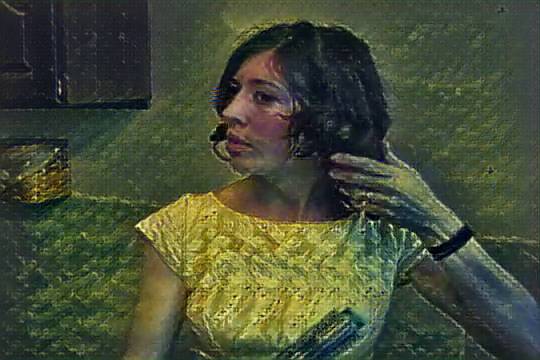}}
    {\caption{}\label{fig:wreck}}
    }
    }
\end{subfloatrow}
}{\caption{Example of visual style transfer applied to the training videos in the MSVD. \ref{fig:or} is an original image from the video. \ref{fig:muse} is the image in the style of the Picasso's painting 'La Muse'. \ref{fig:rain} is the image in the style of the Afremov's painting 'Rain Princess'. \ref{fig:scream} is the image in the style of the Munch's painting 'The Scream'. \ref{fig:udnie} is the image in the style of the Picabia's painting 'Udnie'. \ref{fig:wave} is the image in the style of the Hokusai's painting 'The Great Wave off Kanagawa'. \ref{fig:wreck} is the image in the style of the Turner's painting 'The Wreck of a Transport Ship'.}\label{fig:style}}
\end{figure*}

In addition, the videos have been transformed in the style of some artistic paintings. This was motivated by the fact that the strategy to perform data augmentation via style transfer has been found beneficial for many Computer Vision tasks \cite{jackson2018style}. In this study, the effectiveness of style augmentation for NLVD is investigated. In particular, the approach of \cite{engstrom2016faststyletransfer} has been adopted to transform the videos directly. The applied approach builds on the style transformation network in \cite{johnson2016perceptual} and uses the instance normalization proposed in \cite{vedaldi2016instance} instead of batch normalization. The artistic styles selected are those of Pablo Picasso's 'La Muse', Leonid Afremov's 'Rain Princess', Edvard Munch's 'The Scream', Francis Picabia's 'Udnie', Katsushika Hokusai's 'The Great Wave off Kanagawa', and William Turner's 'The Wreck of a Transport Ship'. Some examples are reported in \figname~\ref{fig:style}.

The applied alterations, either classical or artistic, do not modify the semantic content of the video. In fact, when selecting the transformations, those that would have affected the semantic content of the video have not been considered. For example, cropping, which is a typically applied visual data augmentation strategy, has not been considered to avoid the risk of cropping out something described in the caption. 

\subsection{Data Cleansing to Study the Role of the Textual Input}\label{ssec:text_cleaning}
Providing high-quality training sentences to NLVD models is critical to achieving good performance. Among the available datasets for studying NLVD, in this study the popular MSVD \cite{guadarrama2013youtube2text} is adopted. 
This is the English portion of the dataset presented by Chen and Dolan \cite{chen2011collecting} for paraphrase evaluation. The authors asked AMT workers to describe with a complete, grammatically-correct sentence a short segment of various video clips from YouTube, focussing on the main actor and action depicted. The annotators had the option to watch the entire clip or only the segment to describe, with or without the audio, and could also choose the language for the description. In case English was not the native language of the annotator, the suggestion was given to use the Google translation service. These aspects made possible the collection of low-quality descriptions. Therefore, the authors organized the annotation process in two tasks to describe the same videos. Each annotator performed the first task. According to the quality of the English descriptions provided during the first task, the authors manually granted the best annotators with the access to the second task. Finally, however, the resulting dataset collected the descriptions from both the tasks: $\sim$50k from the first task and $\sim$30k from the second task.

\begin{figure}[]
\centering
 \includegraphics[width = \columnwidth]{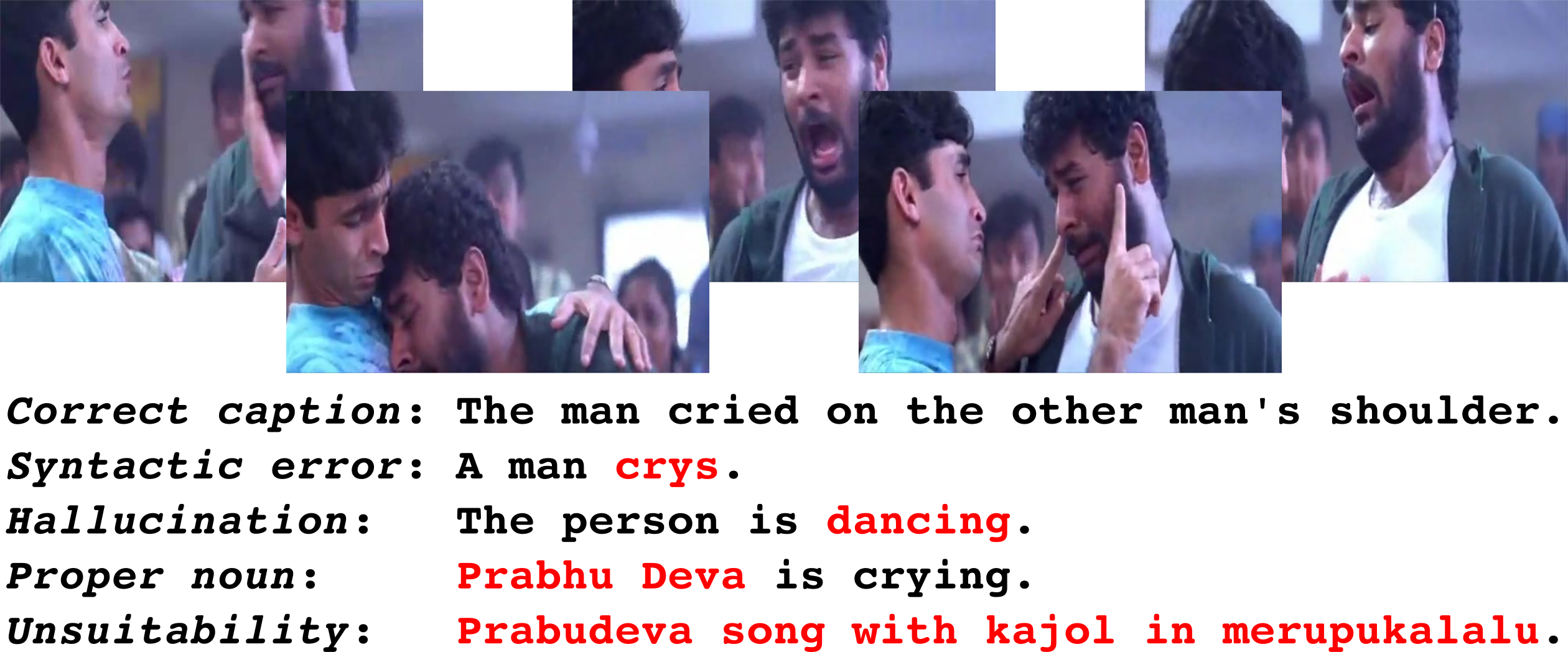}
\caption{Examples of captions with errors associated with a video in the MSVD.}
\label{fig:corrupted_ex}
\end{figure}

Despite the instructions and the quality assessment procedure, the English portion of the MSVD contains syntactically and semantically incorrect descriptions. An example is reported in \figname~\ref{fig:corrupted_ex}. Therefore, for this work, a task involving 21 users has been prepared, in which it has been asked to the users to check and correct all the captions of the MSVD. Note that simply removing the sentences with errors would have reduced the performance, as can be observed from \TABLEname~\ref{tab:n_caption}. For this reason, it has been preferred to amend the errors. Each caption correction has been double checked. For the task, four types of errors have been defined, and the annotators had to find and correct them. The types of errors, ranked based on their severity, are:
\begin{itemize}
\item[1.] unsuitability, \ie the sentence has no meaning, is ill-formulated, or in general, does not respect the instructions given in the original task of \cite{chen2011collecting}. These sentences have been replaced with other correct descriptions of the same video.
\item[2.] hallucination, \ie the sentence describes actors or actions or objects that do not appear in the video. These errors have been corrected double-checking the video.
\item[3.] syntactic, \ie the sentence contains a grammatical error or a typo. These errors have been amended.
\item[4.] proper noun, \ie the sentence contains a proper noun, which cannot be inferred by the video but comes from the experience of the annotator. The proper nouns have been removed or replaced with a common one.
\end{itemize} 

\begin{table}[]
\centering
\caption{BEDDS model performance on the MSVD depending on the number of training captions per video. $B_4$ stands for BLEU\textsubscript{4}, $R_L$ for ROUGE\textsubscript{L}, $M$ for METEOR, and $C$ for CIDEr. Bold indicates the best performance. }
\label{tab:n_caption}
\resizebox{\textwidth}{!}{%
\begin{tabular}{|c|c|cccc|}
\hline
\begin{tabular}[c]{@{}c@{}}Training captions\\ per video\end{tabular} & Training samples & $B_4$ & $R_L$ & $M$ & $C$ \\ \hline 
1 & 1200 & 10.9 & 39.1 & 14.7 & 84.5 \\ \hline
5 & 6000 & 29.8 & 56.7 & 24.1 & \textbf{84.7} \\ \hline
10 & 12000 & 35.3 & 61.2 & 26.9 & 78.9 \\ \hline
$\sim$43 & 49158 & \textbf{45.1} & \textbf{69.4} & \textbf{32.9} & 70.0 \\ \hline
\end{tabular}%
}
\end{table}

In the test subset, the annotators labelled the errors in addition to correcting them. In case of multiple errors, the annotators labelled the caption giving priority to the most severe type of error. From this process, it emerged that the 24.62\% of the captions in the test set contained one or more of the errors just described. The 49.20\% of them had syntactical errors, 27.10\% were unsuitable descriptions for the associated video, the 12.18\% contained hallucinations, and the 11.52\% proper nouns.

To gain insights into the MSVD, both the original one from \cite{chen2011collecting, guadarrama2013youtube2text} and the one obtained for this study, referred to as MSVD-v2, the average human performance has been measured as follows. For each video, a ground truth sentence has been considered the predicted description, and the performance scores have been calculated similarly to what done for the automatic NLVD models. This procedure has been repeated 23 times since each video in the test subset of the MSVD is associated with at least 23 captions. Finally, the mean and standard deviation of the scores have been calculated. The human performance changes after checking the text part of the dataset as reported in \TABLEname~\ref{tab:human}. In particular, the mean value increases for all the scores and the variance decreases. This is no surprise considering the high number of detected and amended errors. In fact, the considered metrics are based on the similarity of words and groups of words in the compared sentences, and the dissimilarity due to the errors has been removed (or significantly reduced.) The values of the scores are not perfect because of the natural diversity in the possible ways to describe each video.

The obtained MSVD-v2 dataset is available online\footnote{\scriptsize {\fontfamily{qcr}\selectfont http://sira.diei.unipg.it/supplementary/input4nlvd2018/}}.
\begin{table}[]
\centering
\caption{Human performance on the MSVD in its original version from \cite{chen2011collecting, guadarrama2013youtube2text} and checked version from this work, MSVD-v2. $B_4$ stands for BLEU\textsubscript{4}, $R_L$ for ROUGE\textsubscript{L}, $M$ for METEOR, and $C$ for CIDEr. Bold indicates the best performance. }
\label{tab:human}
\resizebox{\textwidth}{!}{
\begin{tabular}{c|cccc|}
\cline{2-5}
 & $B_4$ & $R_L$ & $M$ & $C$ \\ 
\hline
\multicolumn{1}{|c|}{MSVD} & 60.10$\pm$6.49 & 77.52$\pm$5.27 & 43.61$\pm$3.59 & 124.30$\pm$20.54 \\ \hline
\multicolumn{1}{|c|}{MSVD-v2} & \textbf{65.40$\pm$4.60} & \textbf{81.93$\pm$2.70} & \textbf{47.45$\pm$1.72} & \textbf{148.64$\pm$16.75}\\
\hline
\end{tabular}}
\end{table}

\section{Experiments and Results}\label{sec:experiments}
In this section, the implementation details of the experimental setup used in this study are reported, and the obtained results are presented and discussed. The BEDDS model described in \ref{ssec:bedds} has been used as the baseline for observing the effects of the visual data augmentation and textual data cleansing preprocessing steps.

\subsection{Implementation Details}
\subsubsection{Architecture Details}
The dimension of the hidden state of the Encoder LSTM and GRU and the Decoder GRU has been set to 1000. When used, the learnt WE maps the words in a \num{300}-dimensional space, and the VE maps the feature vectors in a \num{500}-dimensional space.
The vocabulary $D$ has been built using the training and validation subsets of either the MSVD and MSVD-v2 datasets. In the first case, it consists of \num{10160} words, in the second case, of \num{6428} words.

For the training, the Stochastic Gradient Descent algorithm has been employed, with learning rate set to \num{0.1} and kept constant. The batch size has been set to 64 samples. As the early stopping criterion, the METEOR score on the validation set has been used (similar to what done, \eg in \cite{xu2015show, wang2018reconstruction}.) In particular, the training has been stopped if the value of the METEOR score did not increase for ten consecutive epochs. In the test phase, the best model in terms of the METEOR score has been used. On average, the training ends in $\sim$40 epochs for the models trained the original dataset, $\sim$20 for the style augmented dataset, and $\sim$15 for the classically augmented dataset. This resulted respectively in $\sim$3h, $\sim$24h, and $\sim$48h for training the PyTorch implementation of the models on an NVIDIA Titan XP graphic card.

\subsubsection{Visual Data Augmentation Details}\label{sssec:aug_detail}
Additional to the transformations described in \ref{tab:augmentation}, in the test phase only it has been applied vertical flipping and contrast reduction and enhancement, with multiplicative factors \num{2} and \num{0.5} respectively.
Apart from style transfer, vertical flipping and greyscale conversion, for all the applied transformation a parameter can be set to vary their severity. Different values of the parameters have been chosen for the transformations to the videos in the training set, and others for the tranformations to the test set only. In particular:
\begin{itemize}
\item The kernel size $\rho$ of the gaussian blur has been set to \num{12}, \num{15}, and \num{17} in training phase, and \num{5}, \num{7}, \num{10}, and \num{20} in test phase only.
\item The ratio between the top width $w_{top}$ and the bottom width $w_{bottom}$ of the image for the keystone distortion has been set to \num{5/2}, \num{3}, \num{2/5}, and \num{1/3} in the training phase, and \num{3/2}, \num{2}, \num{2/3}, and \num{1/2} in the test phase only. 
\item The enhancement and reduction factors for the brightness alteration have been set respectively to \num{2} and \num{0.2} in the training phase, and to \num{5}, \num{7}, and \num{0.5}, \num{0.7} in the test phase only. 
\item The probability $p$ that a pixel is affected by the Salt \& Pepper noise has been set to \num{0.01}, \num{0.05}, and \num{0.1} in the training phase, and to \num{0.5}, \num{0.7} in the test phase only.
\end{itemize}
\subsection{Results}
\subsubsection{Effects of Visual Data Augmentation}
The performance of the BEDDS model trained on the original training videos of the MSVD has been evaluated on the MSVD original test videos, and on the same videos altered as explained in \ref{sssec:aug_detail}, to evaluate its generalization capability with respect to different visual conditions. The performance decreases proportionally with the intensity of the various transformations applied. This indicates that the model lacks robustness to unseen appearances of the scene.

The BEDDS model has been trained also on an augmented MSVD, obtained as explained in \ref{ssec:visual_DA} and \ref{sssec:aug_detail}. The resulting models are referred to as BEDDS-VA, in case of training on classically altered videos, and BEDDS-ST in case of training on style transformed videos. The comparison of the three models on the MSVD dataset is reported in \TABLEname~\ref{tab:augmentation} and in \TABLEname~\ref{tab:style}. 

When tested on the original test videos of the MSVD, the performance the BEDDS-VA model is inferior to that of the BEDDS model trained on the original training videos only. However, on the test videos altered with the same transformations as in the training set, the BEDDS-VA model outperforms the BEDDS model in terms of all the metrics. Also for the BEDDS-VA model, the performance decreases proportionally with the intensity of the various transformations applied, but the performance drop is smaller than that of the BEDDS model trained on the original videos only. 
On test videos altered with unseen transformations, including style transfer, the BEDDS-VA model outperforms the BEDDS model in the majority of cases. This is particularly true for the performance in terms of the CIDEr metric, which is the one that by design better captures the human consensus on the quality of image descriptions. The cases in which the performance of the BEDDS model are comparable or superior to that of the BEDDS-VA model are those of transformations that do not significantly alter the appearance of the video, such as vertical flipping and small keystone distortion. 
This suggests that the BEDDS model is biased on the appearance of the training videos.

The BEDDS-ST model outperforms the BEDDS and BEDDS-VA models when tested on the style transformed videos and on severe Salt \& Pepper noise alteration with probability of altered pixel set equals to $0.07$. However, in the majority of the other cases, its performance is inferior to that of the other models. This suggests that, different from other Computer Vision tasks as classification under domain shift and depth estimation, performing style transfer for visual data augmentation is not effective for the NLVD task.

\begin{figure*}[]
\centering
\ffigbox[]{%
\begin{subfloatrow}
	\vbox{
	\hbox{
	\hspace{-0.7cm}
    \ffigbox[\FBwidth]
    {\includegraphics[width=0.53\textwidth]{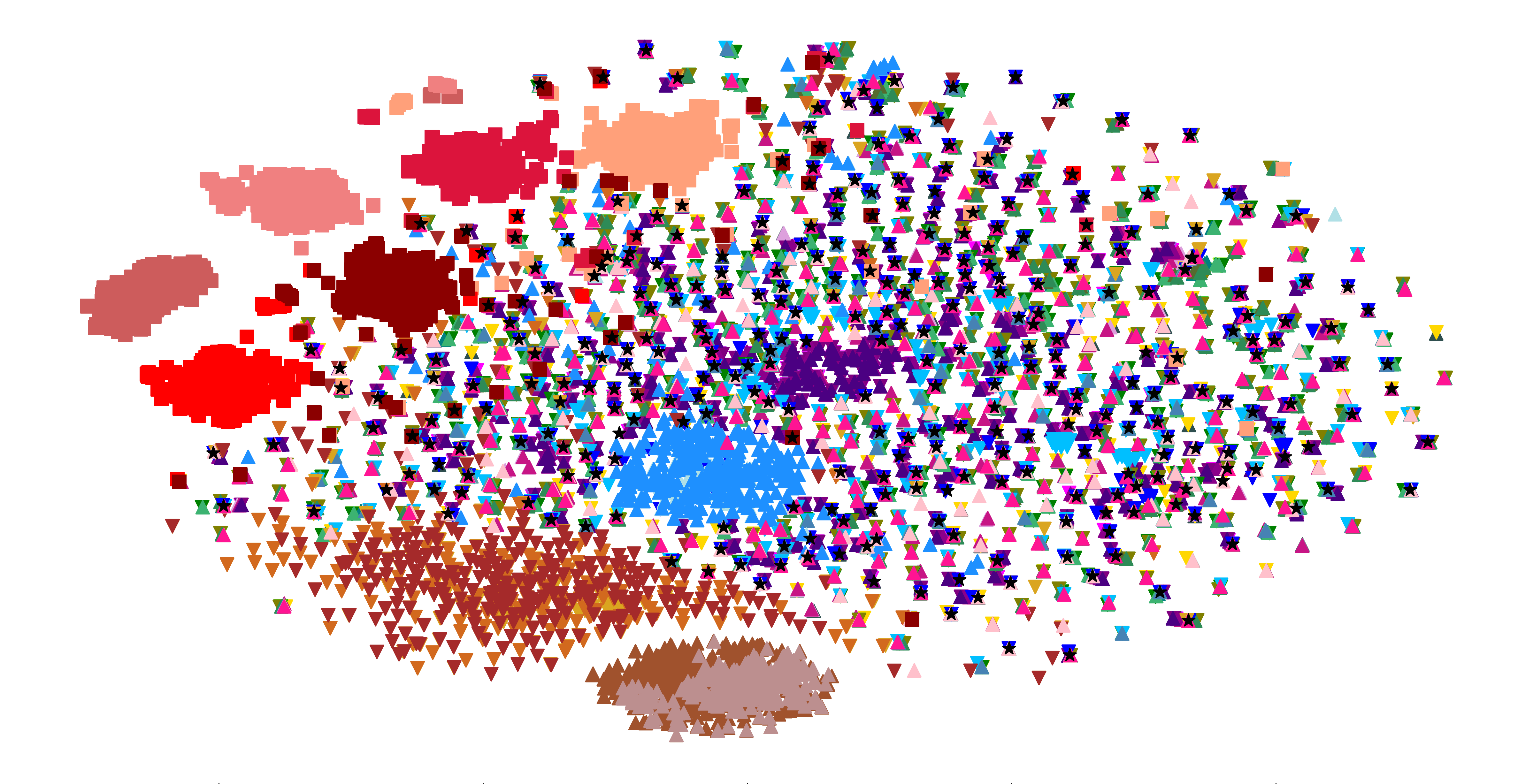}}
    {\caption{}\label{fig:all}}
    \hspace{-0.7cm}
    \ffigbox[\FBwidth]
    {\includegraphics[width=0.53\textwidth]{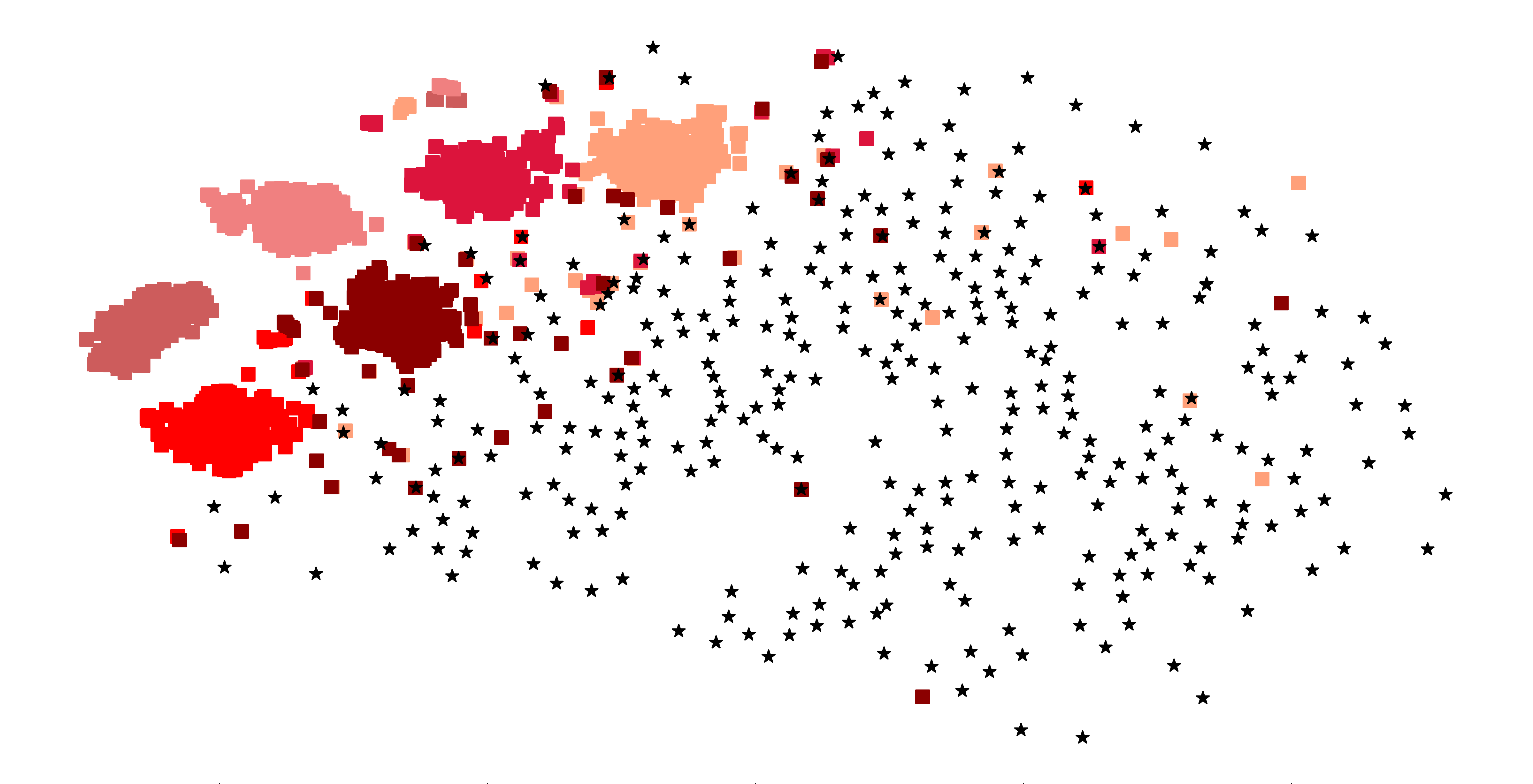}}
    {\caption{}\label{fig:st}}
    }
    \hbox{
    \hspace{-0.7cm}
    \ffigbox[\FBwidth]
    {\includegraphics[width=0.53\textwidth]{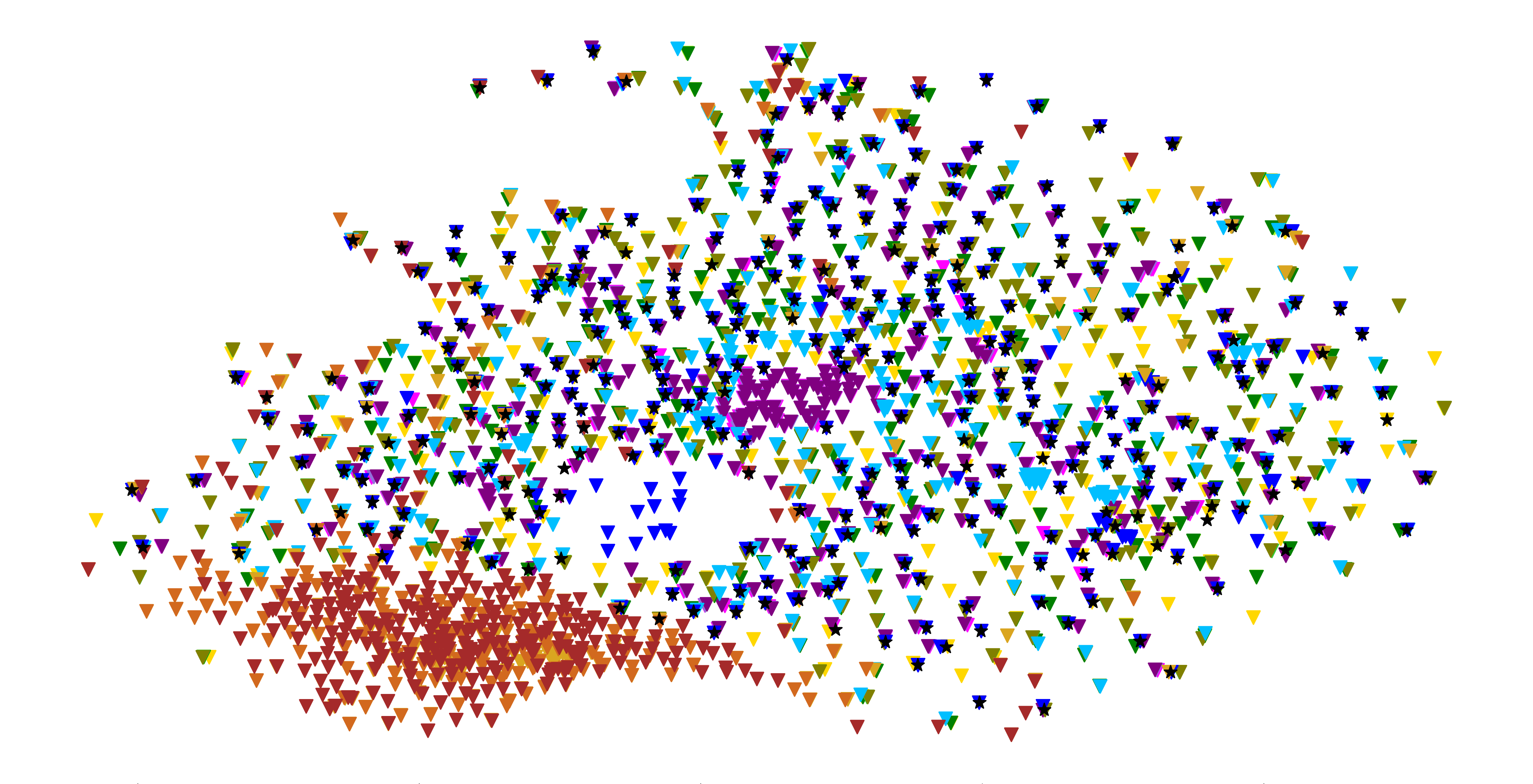}}
    {\caption{}\label{fig:vatr}}
    \hspace{-0.7cm}
    \ffigbox[\FBwidth]
    {\includegraphics[width=0.53\textwidth]{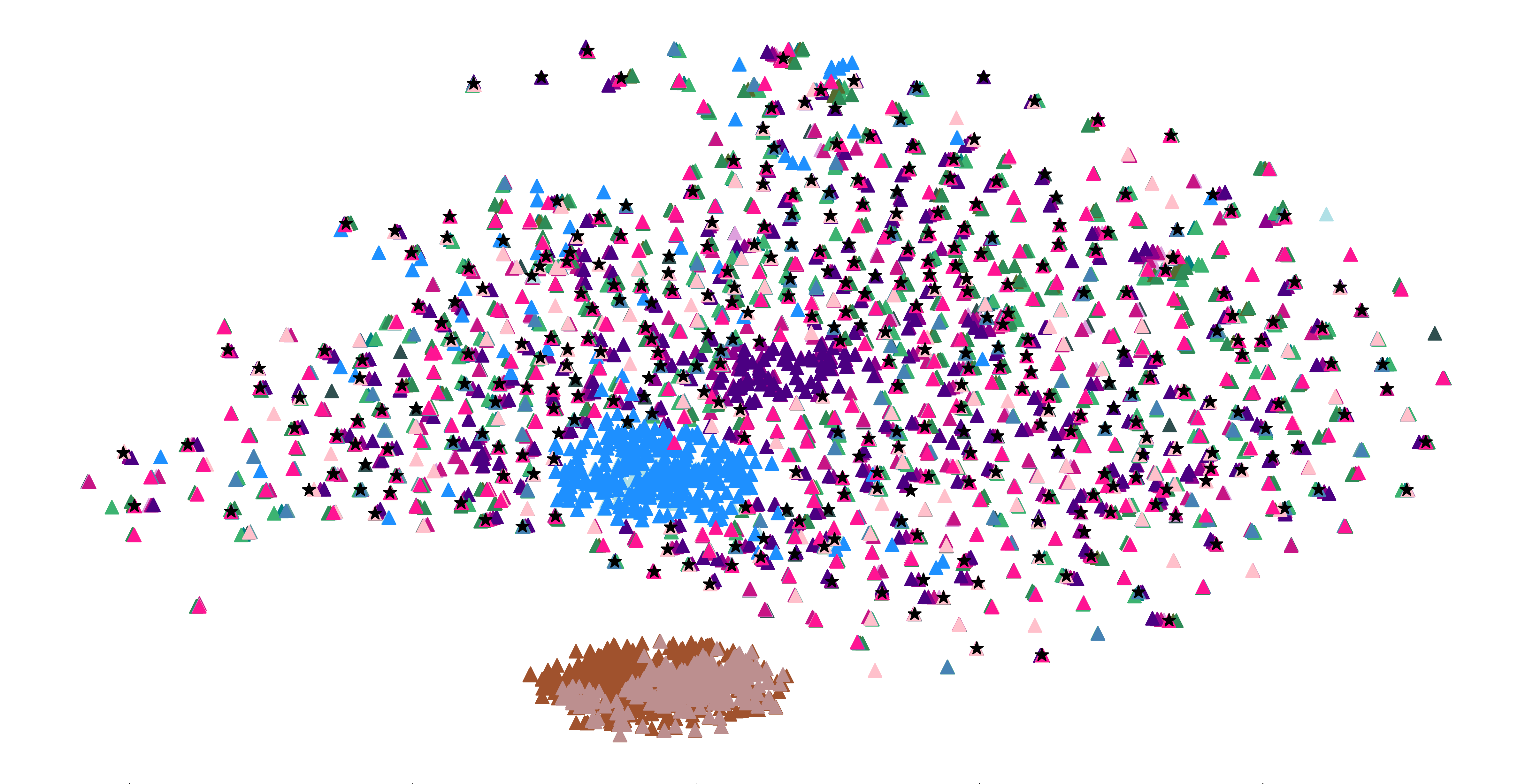}}
    {\caption{}\label{fig:vats}}
	}
    \hbox{
    \ffigbox[\FBwidth]
    {\includegraphics[width=\textwidth]{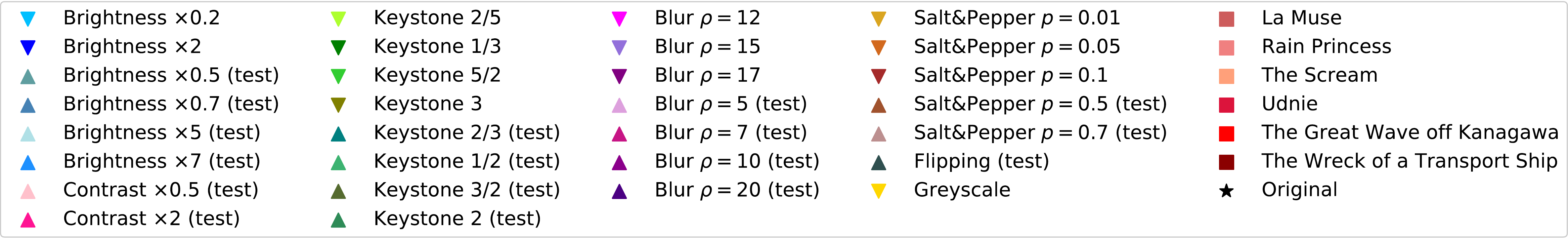}}
    {}}
    }
\end{subfloatrow}}
{\caption{Distribution of the frames in the MSVD dataset, altered with classical alterations and style transfer. \ref{fig:all} contains the points associated to the original frames and to all the altered frames. \ref{fig:st} contains the points associated to the original frames and to the style transformed frames used for training the BEDDS-ST model. \ref{fig:vatr} contains the points associated to the original frames and to the frames altered with the classical alterations in the training set of the BEDDS-VA model. \ref{fig:vats} contains the points associated to the original frames and to the altered frames used only for test.}\label{fig:tsne}}
\end{figure*}

Some intuitions on this behaviour can be gained observing the data distribution obtained via the \emph{t-SNE} analysis \cite{maaten2008visualizing} depicted in \figurename~\ref{fig:tsne}. The points represent the \emph{ResNet50} features extracted from the fifth frame of each video in the MSVD dataset, both original and altered as described in \ref{ssec:visual_DA}. Recall that the \emph{ResNet50} features capture the appearance of the frames. From the \emph{t-SNE} plots, it can be observed that the style transformed frames form separate clusters, which do not overlap with the other data points. Some of the classically altered frames are grouped together, \eg those altered via Salt \& Pepper noise, severe Gaussian blur, and brightness variation. In such cases, the BEDDS-VA model outperforms the BEDDS model. The transformations that do not severely alter the appearance of the videos result in points that are distributed as those corresponding to the original frames. Therefore, in such cases, the BEDDS model performs comparably or better than the BEDDS-VA model. In applicative scenarios, the same analysis can be performed on videos captured under the specific operating conditions. This can guide the selection of the most appropriate visual transformations to apply to videos to include in the dataset for training or finetuning the NLVD system.

Furthermore, the performance of the BEDDS, BEDDS-VA, and BEDDS-ST models have been tested without retraining on the MSR-VTT dataset. This dataset has characteristics similar to those of the MSVD dataset, in terms of visual quality of the videos and number and quality of the captions, since both contain videos from YouTube with multiple captions per videos, collected via the AMT service. The results of this comparison are reported in \TABLEname~\ref{tab:msrvtt}. The performance of the three models are comparable, and all below the human performance on the same dataset, calculated as done for the MSVD dataset in \ref{ssec:text_cleaning}. 

These results suggest that with the visual data augmentation preprocessing step the model can deal better with appearance changes. However, the recently proposed style augmentation approach results less effective than classical alterations in the context of NLVD.
In addition, the robustness with respect to appearance conditions of specific applications can be further increased by training the NLVD models on videos altered accordingly. 

\begin{table*}[t]
\centering
\caption{Performance of the BEDDS, BEDDS-VA, and BEDDS-ST models on differently altered test videos of the MSVD, used both in training and test phase or in test phase only. $B_4$ stands for BLEU\textsubscript{4}, $R_L$ for ROUGE\textsubscript{L}, $M$ for METEOR, and $C$ for CIDEr. Bold indicates the best performance. }
\label{tab:augmentation}
\resizebox{\textwidth}{!}{%
\begin{tabular}{cc|cccc|cccc|cccc|}
\cline{3-14}
 &  & \multicolumn{4}{c|}{BEDDS} & \multicolumn{4}{c|}{BEDDS-VA} & \multicolumn{4}{c|}{BEDDS-ST} \\ \cline{2-14} 
\multicolumn{1}{c|}{} & \textit{Alteration} & $B_4$ & $R_L$ & $M$ & $C$ & $B_4$ & $R_L$ & $M$ & $C$ & $B_4$ & $R_L$ & $M$ & $C$ \\ \hline \cline{2-14}
\multicolumn{1}{|c|}{\multirow{14}{*}{\begin{sideways} In BEDDS-VA Training and Test Phase \end{sideways}}} & None (\ie Original videos) & \textbf{45.1} & \textbf{69.4} & \textbf{32.9} & \textbf{70.0} & 43.5 & 69.2 & 32.6 & 69.7 & 42.5 & 68.1 & 31.6 & 62.7 \\ \Cline{1pt}{2-14}
\multicolumn{1}{|c|}{} & Greyscale Conversion & 40.4 & 66.6 & 30.7 & 57.8 & \textbf{43.5} & \textbf{69.1} & \textbf{32.0} & \textbf{66.3} & 40.1 & 66.9 & 30.4 & 59.1 \\ \cline{2-14} 
\multicolumn{1}{|c|}{} & Gaussian Blur with $\rho = 12$ & 39.2 & 65.6 & 29.6 & 50.6 & \textbf{41.4} & \textbf{68.6} & \textbf{30.9} & \textbf{64.0} & 37.7 & 64.8 & 28.5 & 46.8 \\ \cline{2-14} 
\multicolumn{1}{|c|}{} & Gaussian Blur with $\rho = 15$ & 36.4 & 64.5 & 28.3 & 44.3 & \textbf{40.5} & \textbf{66.6} & \textbf{30.4} & \textbf{59.6} & 35.0 & 63.4 & 27.3 & 41.1 \\ \cline{2-14} 
\multicolumn{1}{|c|}{} & Gaussian Blur with $\rho = 17$ & 34.8 & 63.3 & 27.4 & 39.3 & \textbf{43.2} & \textbf{68.6} & \textbf{31.7} & \textbf{67.1} & 33.3 & 62.7 & 26.7 & 37.6 \\ \cline{2-14}
\multicolumn{1}{|c|}{} & Keystone Distortion $w_{top}/w_{bottom} = 2/5$ & 40.3 & 67.2 & 30.9 & 59.1 & \textbf{43.2} & \textbf{68.6} & \textbf{31.7} & \textbf{70.2} & 36.0 & 64.0 & 28.3 & 47.5 \\ \cline{2-14} 
\multicolumn{1}{|c|}{} & Keystone Distortion $w_{top}/w_{bottom} = 1/3$ & 38.9 & 66.6 & 30.4 & 56.0 & \textbf{40.0} & \textbf{67.7} & \textbf{30.8} & \textbf{61.9} & 34.1 & 63.3 & 27.1 & 40.4 \\ \cline{2-14} 
\multicolumn{1}{|c|}{} & Keystone Distortion $w_{top}/w_{bottom} = 5/2$ & 39.1 & 66.7 & 30.2 & 59.5 & \textbf{41.5} & \textbf{67.6} & \textbf{31.1} & \textbf{66.3} & 37.3 & 64.5 & 28.3 & 50.0 \\ \cline{2-14} 
\multicolumn{1}{|c|}{} & Keystone Distortion $w_{top}/w_{bottom} = 3$ & 36.3 & 65.3 & 28.8 & 48.8 & \textbf{40.7} & \textbf{66.9} & \textbf{29.4} & \textbf{59.7} & 34.3 & 62.7 & 26.7 & 43.7 \\ \cline{2-14}
\multicolumn{1}{|c|}{} & Brightness Reduction $\times0.2$ & 38.7 & 64.9 & 29.0 & 53.8 & \textbf{42.1} & \textbf{68.3} & \textbf{31.2} & \textbf{64.6} & 38.0 & 65.0 & 28.8 & 53.9 \\ \cline{2-14} 
\multicolumn{1}{|c|}{} & Brightness Enhancement $\times2$ & 39.2 & 67.1 & 30.5 & 57.3 & \textbf{42.0} & \textbf{67.8} & \textbf{31.2} & \textbf{64.0} & 37.3 & 65.3 & 29.2 & 54.0 \\ \cline{2-14}
\multicolumn{1}{|c|}{} & Salt \& Pepper noise with $p=0.01$ & 26.5 & 59.9 & 24.3 & 36.3 & \textbf{40.4} & \textbf{66.6} & \textbf{31.0} & \textbf{62.6} & 33.7 & 62.8 & 27.1 & 43.2 \\ \cline{2-14}
\multicolumn{1}{|c|}{} & Salt \& Pepper noise with $p=0.05$ & 22.7 & 58.7 & 23.1 & 26.4 & \textbf{39.2} & \textbf{65.9} & \textbf{30.0} & \textbf{54.6} & 30.5 & 61.5 & 25.6 & 32.8 \\ \cline{2-14} 
\multicolumn{1}{|c|}{} & Salt \& Pepper noise with $p=0.1$ & 22.9 & 58.7 & 23.1 & 23.8 & \textbf{38.0} & \textbf{65.4} & \textbf{29.4} & \textbf{53.2} & 28.2 & 59.9 & 24.5 & 29.3 \\ \hline
\multicolumn{1}{|c|}{\multirow{17}{*}{\begin{sideways}In Test Phase Only\end{sideways}}} & Vertical Flipping & \textbf{44.5} & \textbf{69.4} & \textbf{32.9} & \textbf{70.3} & 43.4 & 69.1 & 32.3 & 68.3 & 40.6 & 67.2 & 30.8 & 58.6 \\ \cline{2-14} 
\multicolumn{1}{|c|}{} & Gaussian Blur with $\rho = 5$ & \textbf{44.4} & 69.0 & 32.2 & 68.5 & 43.6 & \textbf{69.1} & \textbf{32.4} & \textbf{69.9} & 41.5 & 67.6 & 30.8 & 58.6 \\ \cline{2-14} 
\multicolumn{1}{|c|}{} & Gaussian Blur with $\rho = 7$ & \textbf{43.2} & 68.1 & 31.5 & 62.5 & 42.7 & \textbf{68.7} & \textbf{32.0} & \textbf{67.8} & 40.3 & 67.0 & 30.6 & 56.2 \\ \cline{2-14}
\multicolumn{1}{|c|}{} & Gaussian Blur with $\rho = 10$ & 41.4 & 67.0 & 30.7 & 57.3 & \textbf{42.5} & \textbf{68.3} & \textbf{31.9} & \textbf{66.5} & 39.6 & 65.9 & 29.4 & 51.7 \\ \cline{2-14} 
\multicolumn{1}{|c|}{} & Gaussian Blur with $\rho = 20$ & 32.2 & 61.7 & 26.2 & 32.9 & \textbf{38.4} & \textbf{65.4} & \textbf{29.3} & \textbf{54.9} & 31.7 & 61.8 & 25.8 & 32.0 \\ \cline{2-14}
\multicolumn{1}{|c|}{} & Keystone Distortion $w_{top}/w_{bottom} = 2/3$ & \textbf{44.3} & 69.1 & \textbf{32.6} & 67.8 & 43.6 & \textbf{69.4} & 32.3 & \textbf{70.8} & 42.7 & 67.8 & 31.6 & 62.0 \\ \cline{2-14} 
\multicolumn{1}{|c|}{} & Keystone Distortion $w_{top}/w_{bottom} = 1/2$ & \textbf{42.8} & \textbf{68.5} & \textbf{31.7} & 64.5 & 40.9 & 67.8 & 31.6 & \textbf{71.5} & 39.1 & 66.0 & 29.9 & 53.9 \\ \cline{2-14}
\multicolumn{1}{|c|}{} & Keystone Distortion $w_{top}/w_{bottom} = 3/2$ & \textbf{45.4} & \textbf{69.5} & \textbf{32.6} & \textbf{69.8} & 43.5 & \textbf{69.5}& 32.5 & \textbf{69.8} & 42.4 & 67.9 & 31.5 & 64.9 \\ \cline{2-14} 
\multicolumn{1}{|c|}{} & Keystone Distortion $w_{top}/w_{bottom} = 2$ & 41.4 & 67.9 & 31.3 & 64.3 & \textbf{43.1} & \textbf{68.7} & \textbf{32.1} & \textbf{69.8} & 39.7 & 65.8 & 29.6 & 57.7 \\ \cline{2-14}
\multicolumn{1}{|c|}{} & Brightness Reduction $\times0.5$ & \textbf{44.2} & \textbf{69.2} & 32.3 & 68.2 & 43.4 & 69.1 & \textbf{32.5} & \textbf{70.7} & 41.4 & 67.8 & 31.0 & 61.8 \\ \cline{2-14}
\multicolumn{1}{|c|}{} & Brightness Reduction $\times0.7$ & 45.1 & \textbf{69.4} & \textbf{32.8} & 71.6 & \textbf{46.6} & 69.2 & 32.6 & \textbf{74.2} & 41.4 & 67.9 & 31.2 & 61.6 \\ \cline{2-14} 
\multicolumn{1}{|c|}{} & Brightness Enhancement $\times5$ & 24.0 & 57.7 & 23.1 & 29.1 & \textbf{27.1} & \textbf{59.2} & \textbf{24.4} & \textbf{33.3} & 25.2 & 57.4 & 23.2 & 26.8 \\ \cline{2-14} 
\multicolumn{1}{|c|}{} & Brightness Enhancement $\times7$ & 19.4 & \textbf{55.2} & 21.2 & 18.2 & \textbf{21.5} & \textbf{55.2} & \textbf{21.8} & \textbf{21.2} & 18.9 & 53.4 & 20.2 & 16.9 \\ \cline{2-14} 
\multicolumn{1}{|c|}{} & Salt \& Pepper noise with $p=0.5$ & 10.0 & \textbf{52.7} & 18.6 & 2.8 & \textbf{15.4} & 52.2 & 17.8 & \textbf{7.8} & 12.8 & 52.6 & \textbf{19.0} & 3.2 \\ \cline{2-14} 
\multicolumn{1}{|c|}{} & Salt \& Pepper noise with $p=0.7$ & 8.4 & 51.9 & 18.6 & 1.8 & 9.7 & 49.0 & 14.7 & 2.0 & \textbf{10.4} & \textbf{53.9} & \textbf{21.0} & \textbf{2.1} \\ \cline{2-14} 
\multicolumn{1}{|c|}{} & Contrast Reduction $\times0.5$ & \textbf{44.0} & 68.5 & 31.8 & 64.7 & 42.8 & \textbf{69.0} & \textbf{31.9} & \textbf{68.1} & 42.3 & 67.8 & 31.2 & 62.4 \\ \cline{2-14} 
\multicolumn{1}{|c|}{} & Contrast Enhancement $\times2$ & 41.5 & 67.5 & 30.9 & 60.9 & \textbf{41.9} & \textbf{68.6} & \textbf{31.6} & \textbf{65.6} & 38.7 & 66.1 & 29.7 & 57.7 \\ \hline
\end{tabular}%
}
\end{table*}

\begin{table*}[]
\centering
\caption{Performance of the BEDDS, BEDDS-VA, and BEDDS-ST models on the  test videos of the MSVD, transformed in different artistic styles. $B_4$ stands for BLEU\textsubscript{4}, $R_L$ for ROUGE\textsubscript{L}, $M$ for METEOR, and $C$ for CIDEr. Bold indicates the best performance. }
\label{tab:style}
\resizebox{\textwidth}{!}{%
\begin{tabular}{c|cccc|cccc|cccc|}
\cline{2-13}
 & \multicolumn{4}{c}{BEDDS} & \multicolumn{4}{c|}{BEDDS-VA} & \multicolumn{4}{c|}{BEDDS-ST} \\ \hline
\multicolumn{1}{|c|}{\textit{Style}} &$B_4$ & $R_L$ & $M$ & $C$ & $B_4$ & $R_L$ & $M$ & $C$ & $B_4$ & $R_L$ & $M$ & $C$ \\ \Cline{1pt}{1-13}
\multicolumn{1}{|c|}{Original videos} & \textbf{45.1} & \textbf{69.4} & \textbf{32.9} & \textbf{70.0} & 43.5 & 69.2 & 32.6 & 69.7 & 42.5 & 68.1 & 31.6 & 62.7 \\ \Cline{1pt}{1-13} 
\multicolumn{1}{|c|}{Picasso's La Muse} & 10.3 & 43.4 & 16.4 & 7.6 & 13.6 & 51.3 & 19.2 & 7.2 & \textbf{31.9} & \textbf{61.8} & \textbf{26.5} & \textbf{38.9} \\ \hline
\multicolumn{1}{|c|}{Afremov's Rain Princess} & 18.5 & 51.1 & 19.9 & 20.6 & 13.9 & 48.2 & 19.2 & 12.4 & \textbf{34.8} & \textbf{63.3} & \textbf{26.8} & \textbf{44.0} \\\hline
\multicolumn{1}{|c|}{Munch's The Scream} & 27.9 & 60.5 & 25.5 & 36.9 & 27.8 & 60.5 & 25.4 & 38.5 & \textbf{38.9} & \textbf{65.8} & \textbf{29.6} & \textbf{54.4} \\\hline
\multicolumn{1}{|c|}{Picabia's Udnie} & 23.0 & 57.2 & 23.1 & 21.6 & 23.0 & 57.8 & 23.2 & 21.4 & \textbf{35.5} & \textbf{64.2} & \textbf{27.9} & \textbf{48.0} \\\hline
\multicolumn{1}{|c|}{Hokusai's The Great Wave off Kanagawa} & 22.9 & 57.9 & 22.8 & 24.6 & 24.2 & 58.4 & 24.1 & 26.9 & \textbf{38.8} & \textbf{65.5} & \textbf{29.1} & \textbf{50.8} \\\hline
\multicolumn{1}{|c|}{Turner's The Wreck of a Transport Ship} & 25.4 & 59.3 & 23.6 & 30.2 & 26.6 & 60.5 & 24.8 & 34.8 & \textbf{38.2} & \textbf{65.3} & \textbf{28.9} & \textbf{52.8} \\ \hline
\end{tabular}%
}
\end{table*}

\begin{table}[]
\centering
\resizebox{\textwidth}{!}{%
\begin{tabular}{c|cccc|}
\cline{2-5}
\textit{}                      & $B_4$          & $R_L$          & $M$            & $C$            \\ \hline
\multicolumn{1}{|c|}{BEDDS}    & 16.9           & 42.7           & 16.3           & \textbf{9.6}   \\ \hline
\multicolumn{1}{|c|}{BEDDS-VA} & 16.9           & 42.2           & 16.0           & 9.0            \\ \hline
\multicolumn{1}{|c|}{BEDDS-ST} & 16.9           & 42.1           & 16.0           & 8.6            \\ \hline
\multicolumn{1}{|c|}{BEDDS-TC} & \textbf{17.5}  & \textbf{43.0}  & \textbf{16.5}  & 9.3            \\ \hline
\multicolumn{1}{|c|}{Humans}   & 23.4 $\pm$ 3.6 & 44.7 $\pm$ 1.2 & 23.5 $\pm$ 0.7 & 31.2 $\pm$ 1.6 \\ \hline
\end{tabular}%
}
\caption{Performance of the BEDDS, BEDDS-VA, and BEDDS-ST models on the original test videos of the MSR-VTT. $B_4$ stands for BLEU\textsubscript{4}, $R_L$ for ROUGE\textsubscript{L}, $M$ for METEOR, and $C$ for CIDEr. Bold indicates the best performance of the models. For completeness, the human performance are also reported.}
\label{tab:msrvtt}
\end{table}

\subsubsection{Effects of Textual Data Cleansing}
The BEDDS model has been trained on either the MSVD and the MSVD-v2 dataset, obtained as explained in \ref{ssec:text_cleaning}. The resulting model is referred to as BEDDS-TC. Both the variants have been tested on the two datasets. The BEDDS-TC model outperforms the BEDDS model on both datasets in terms of all metrics but CIDEr on the MSVD (69.8 for BEDDS-TC and 70.0 for BEDDS.) The results of this study are reported in \TABLEname~\ref{tab:cleaning}. The same trend can be observed also when testing on the MSR-VTT dataset, as observed from \TABLEname~\ref{tab:msrvtt}.
\begin{table}[]
\centering
\caption{Performance of the BEDDS and BEDDS-TC models on the two versions of the MSVD, original and checked (MSVD-v2). $B_4$ stands for BLEU\textsubscript{4}, $R_L$ for ROUGE\textsubscript{L}, $M$ for METEOR, and $C$ for CIDEr. Bold indicates the best performance. }
\label{tab:cleaning}
\resizebox{\textwidth}{!}{%
\begin{tabular}{cc|cccc|}
\cline{3-6}
 & & {\footnotesize $B_4$} & {\footnotesize $R_L$} & {\footnotesize $M$} & {\footnotesize $C$} \\ \hline
\multicolumn{1}{|c|}{\multirow{2}{*}{{\footnotesize MSVD}}} & {\footnotesize BEDDS} & {\footnotesize 45.1} & {\footnotesize 69.4} & {\footnotesize 32.9} & {\footnotesize \textbf{70.0}} \\ \cline{2-6} 
\multicolumn{1}{|c|}{} & {\footnotesize BEDDS-TC} & {\footnotesize \textbf{45.8}} & {\footnotesize \textbf{70.1}} & {\footnotesize \textbf{33.1}} & {\footnotesize 69.8} \\ \hline
\multicolumn{1}{|c|}{\multirow{2}{*}{{\footnotesize MSVD-v2}}} & {\footnotesize BEDDS} & {\footnotesize 44.6} & {\footnotesize 69.2} & {\footnotesize 32.6} & {\footnotesize 68.7} \\ \cline{2-6} 
\multicolumn{1}{|c|}{} & {\footnotesize BEDDS-TC} & {\footnotesize \textbf{45.5}} & {\footnotesize \textbf{70.0}} & {\footnotesize \textbf{33.1}} & {\footnotesize \textbf{79.5}} \\ \hline
\end{tabular}%
}
\end{table}

\begin{figure}[]
\centering
\includegraphics[width =\columnwidth]{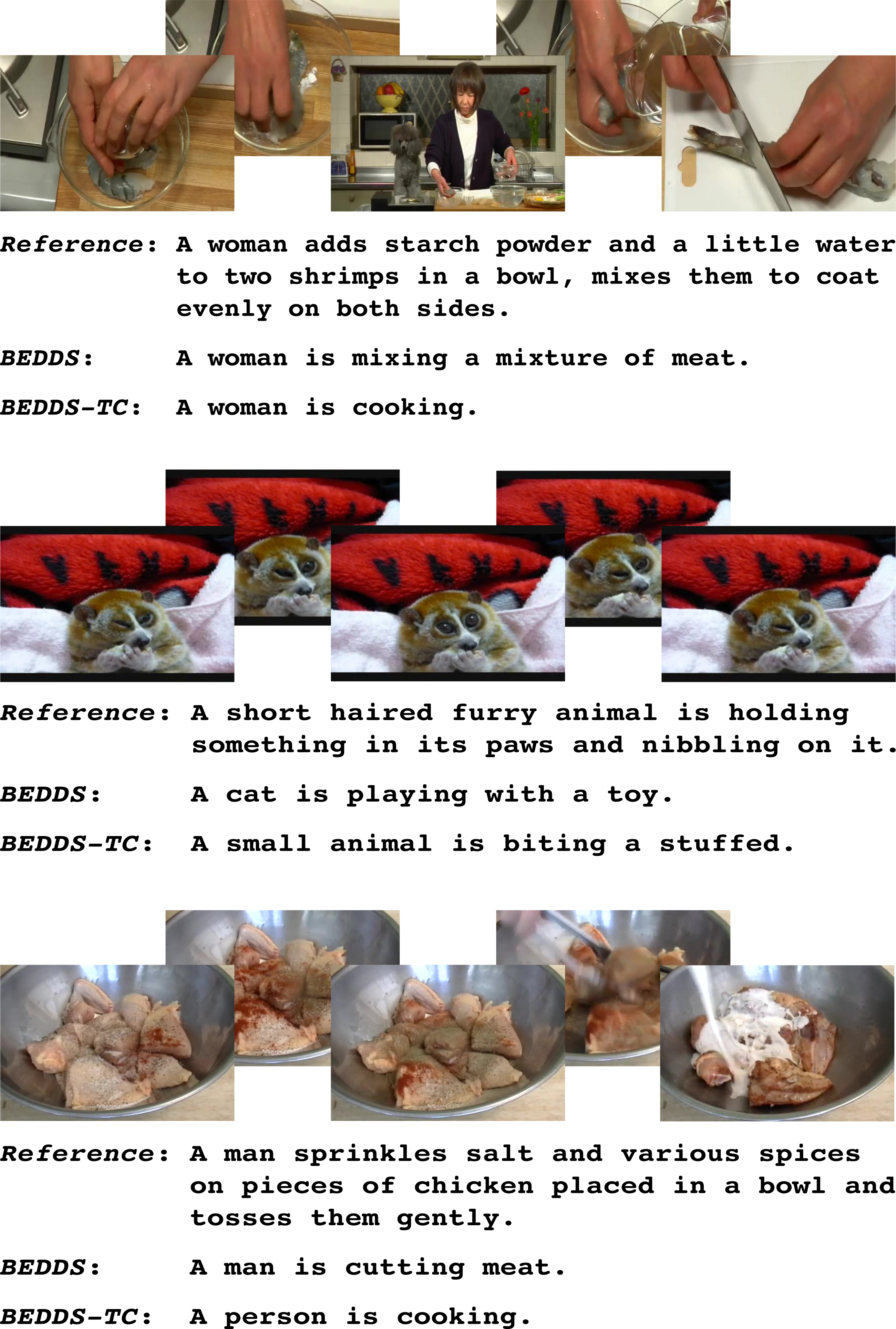}
\caption{Exemplar captions produced by the BEDDS model, which was trained on the original MSVD, and the BEDDS-TC model, which was trained on the MSVD-v2 dataset. }
\label{fig:examples}
\end{figure}

\begin{figure}[]
\centering
\includegraphics[width =\columnwidth]{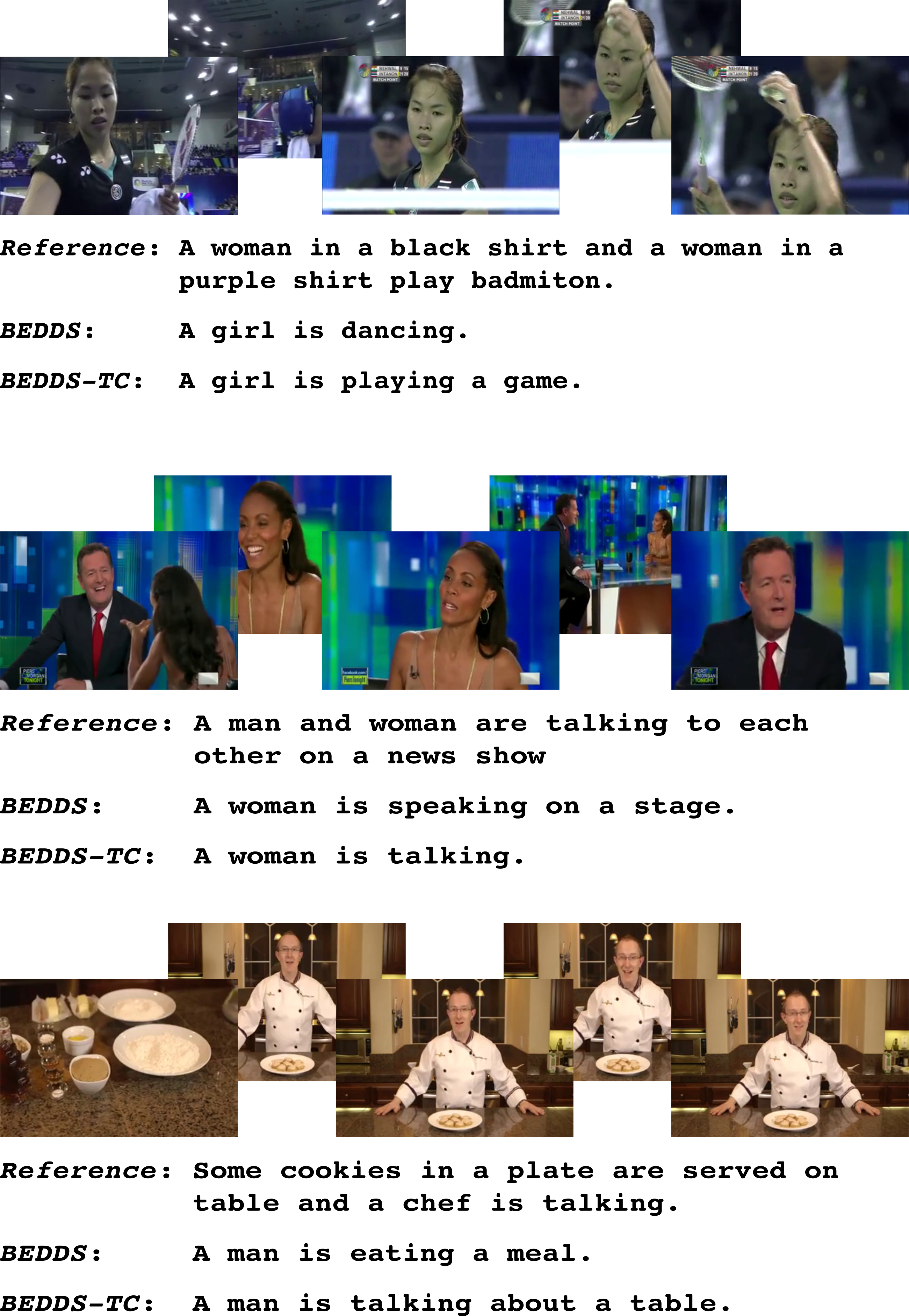}
\caption{Exemplar captions produced by the BEDDS model and the BEDDS-TC model on the MSR-VTT dataset. }
\label{fig:examplesmsrvtt}
\end{figure}

Considering only the performance gain obtained in terms of evaluation metrics is limiting and can be misleading for investigating the effects of training with high-quality textual data. 
Therefore, the descriptions produced by the two models have been compared further, from a qualitative point of view. The complete corpus of results is available online\footnote{{\scriptsize \fontfamily{qcr}\selectfont http://sira.diei.unipg.it/supplementary/input4nlvd2018/}}. As expected, there are cases where one model outputs a correct description while the other a completely wrong one. Nevertheless, both the BEDDS-TC and the BEDDS models produce correct detailed descriptions for the same videos. It is interesting to focus on the cases where the BEDDS model outputs an erroneous detailed description. Some examples are reported in \figurename~\ref{fig:examples} for the MSVD dataset, and in \figurename~\ref{fig:examplesmsrvtt} for the MSR-VTT dataset. In such cases, the descriptions from the BEDDS-TC model are more generic but still correct. However, metrics based on \emph{n}-gram similarity rather than semantic consistency, like those used in the NLVD evaluation, cannot properly capture this aspect. In addition, synonyms and hypernyms can be penalized \cite{kilickaya2016re}. This can explain the little performance gain achieved with textual data cleansing in terms of such metrics.
In the sight of this and of the considerations in \ref{ssec:text_cleaning} on the syntactic and semantic errors in the MSVD, we believe that using the MSVD-v2 dataset to train and test the NLVD algorithms is reasonable because it contains better quality ground truth captions. This is confirmed by the average human performance estimation on the MSVD and MSVD-v2 datasets.  As mentioned in \ref{ssec:text_cleaning}, its mean value is higher on the amended dataset, and the variance is smaller. Neither human performance can be perfect for this task, due to its intrinsic subjectivity. However, the improved performance after the textual data cleansing suggests that the MSVD-v2 dataset represents a more reliable benchmark than the MSVD for the NLVD task.
Finally, the comparison of the performance on original and the amended datasets highlights the importance of the consistency of the textual component when designing an NLVD system. 

\section{Conclusion}\label{sec:conclusion}
In this work, it has been presented a study to evaluate the performance of NLVD systems in case the video input dataset is augmented with transformed video derived from the original ones applying common transformations. For this purpose, extensive studies have been performed on the benchmark MSVD dataset and on a refined version specifically amended for this study (the MSVD-v2 dataset.) 
The experiments have been carried out using a simple yet effective NLVD encoder-decoder architecture. 

The results of the analysis reveal that the visual data augmentation generally provides improvements in terms of robustness to appearance changes. In particular, considering the CIDEr score, which by design correlates with the human judgment on image description, the model trained on the augmented videos obtains an average $+4.5\%$ performance improvement with peaks up to $+22.0\%$ for severe Gaussian blur, when tested on videos altered using a different set of transformations compared to those used in the training set. As expected, this improvement is more significant when the NLVD model is tested on videos altered with the same transformation used in the training set ($+12.7\%$ on average, with peaks up to $+29,4\%$ for severe alterations as keystone distortion and Salt \& Pepper noise.) This suggests that, when applying the NLVD system in a real-world scenario, it is beneficial to train or finetune the system with videos altered according to the visual conditions typical of the specific application. 
In this work, it has been shown that some insights on the utility of the specific input transformations can be gained using a t-SNE analysis. Specifically, the videos altered via transformations that do not severely change the appearance are distributed as the original videos, while those altered with severe transformations (such as  Salt \& Pepper noise, severe Gaussian blur and brightness variation) are grouped in separate clusters. For those latter cases, data augmentation brings to a significant improvement in the performance of the NLVD system. Finally, it was observed that the BEDDS-TC model, trained on the refined MSVD-v2 dataset, provides more generic but correct captions, refelcted in a performance improvement in terms of all the evaluation metrics.

\section*{Acknowledgment}
We gratefully thank the NVIDIA Corporation with the donation of the \emph{Titan XP} GPU used for this research. Our gratitude also goes to the users who volunteered for the MSVD text checking task.

\bibliographystyle{IEEEtran}
\bibliography{\include/bibliography}

\begin{thebibliography}{10}
\providecommand{\url}[1]{#1}
\csname url@rmstyle\endcsname
\providecommand{\newblock}{\relax}
\providecommand{\bibinfo}[2]{#2}
\providecommand\BIBentrySTDinterwordspacing{\spaceskip=0pt\relax}
\providecommand\BIBentryALTinterwordstretchfactor{4}
\providecommand\BIBentryALTinterwordspacing{\spaceskip=\fontdimen2\font plus
\BIBentryALTinterwordstretchfactor\fontdimen3\font minus
  \fontdimen4\font\relax}
\providecommand\BIBforeignlanguage[2]{{%
\expandafter\ifx\csname l@#1\endcsname\relax
\typeout{** WARNING: IEEEtran.bst: No hyphenation pattern has been}%
\typeout{** loaded for the language `#1'. Using the pattern for}%
\typeout{** the default language instead.}%
\else
\language=\csname l@#1\endcsname
\fi
#2}}

\bibitem{cui2018general}
P.~Cui, S.~Liu, and W.~Zhu, ``General knowledge embedded image representation
  learning,'' \emph{IEEE Transactions on Multimedia}, vol.~20, no.~1, pp.
  198--207, 2018.

\bibitem{kofler2014predicting}
C.~Kofler, L.~Yang, M.~Larson, T.~Mei, A.~Hanjalic, and S.~Li, ``Predicting
  failing queries in video search,'' \emph{IEEE Transactions on Multimedia},
  vol.~16, no.~7, pp. 1973--1985, 2014.

\bibitem{xie2014contextual}
H.~Xie, Y.~Zhang, J.~Tan, L.~Guo, and J.~Li, ``Contextual query expansion for
  image retrieval,'' \emph{IEEE Transactions on Multimedia}, vol.~16, no.~4,
  pp. 1104--1114, 2014.

\bibitem{li2017joint}
W.~Li, J.~Joo, H.~Qi, and S.-C. Zhu, ``Joint image-text news topic detection
  and tracking by multimodal topic and-or graph.'' \emph{IEEE Trans.
  Multimedia}, vol.~19, no.~2, pp. 367--381, 2017.

\bibitem{hu2018twitter100k}
Y.~Hu, L.~Zheng, Y.~Yang, and Y.~Huang, ``Twitter100k: A real-world dataset for
  weakly supervised cross-media retrieval,'' \emph{IEEE Transactions on
  Multimedia}, vol.~20, no.~4, pp. 927--938, 2018.

\bibitem{song2018extracting}
H.~Song, X.~Wu, W.~Yu, and Y.~Jia, ``Extracting key segments of videos for
  event detection by learning from web sources,'' \emph{IEEE Transactions on
  Multimedia}, vol.~20, no.~5, pp. 1088--1100, 2018.

\bibitem{yang2018text2video}
X.~Yang, T.~Zhang, and C.~Xu, ``Text2video: An end-to-end learning framework
  for expressing text with videos,'' \emph{IEEE Transactions on Multimedia},
  2018.

\bibitem{baraldi2017recognizing}
L.~Baraldi, C.~Grana, and R.~Cucchiara, ``Recognizing and presenting the
  storytelling video structure with deep multimodal networks,'' \emph{IEEE
  Transactions on Multimedia}, vol.~19, no.~5, pp. 955--968, 2017.

\bibitem{li2018gla}
L.~Li, S.~Tang, Y.~Zhang, L.~Deng, and Q.~Tian, ``Gla: Global--local attention
  for image description,'' \emph{IEEE Transactions on Multimedia}, vol.~20,
  no.~3, pp. 726--737, 2018.

\bibitem{gao2017video}
L.~Gao, Z.~Guo, H.~Zhang, X.~Xu, and H.~T. Shen, ``Video captioning with
  attention-based lstm and semantic consistency,'' \emph{IEEE Transactions on
  Multimedia}, vol.~19, no.~9, pp. 2045--2055, 2017.

\bibitem{dong2018predicting}
J.~Dong, X.~Li, and C.~G. Snoek, ``Predicting visual features from text for
  image and video caption retrieval,'' \emph{IEEE Transactions on Multimedia},
  2018.

\bibitem{vedantam2015cider}
R.~Vedantam, C.~Lawrence~Zitnick, and D.~Parikh, ``C{I}{D}{E}r: Consensus-based
  {I}mage {D}escription {E}valuation,'' in \emph{Proceedings of the IEEE
  Conference on Computer Vision and Pattern Recognition}, 2015, pp. 4566--4575.

\bibitem{anderson2016spice}
P.~Anderson, B.~Fernando, M.~Johnson, and S.~Gould, ``Spice: Semantic
  propositional image caption evaluation,'' in \emph{European Conference on
  Computer Vision}.\hskip 1em plus 0.5em minus 0.4em\relax Springer, 2016, pp.
  382--398.

\bibitem{cascianelli2018full}
S.~Cascianelli, G.~Costante, T.~A. Ciarfuglia, P.~Valigi, and M.~L. Fravolini,
  ``Full-gru natural language video description for service robotics
  applications,'' \emph{IEEE Robotics and Automation Letters}, vol.~3, no.~2,
  pp. 841--848, 2018.

\bibitem{torabi2015using}
A.~Torabi, C.~Pal, H.~Larochelle, and A.~Courville, ``Using descriptive video
  services to create a large data source for video annotation research,''
  \emph{arXiv preprint arXiv:1503.01070}, 2015.

\bibitem{rohrbach2015dataset}
A.~Rohrbach, M.~Rohrbach, N.~Tandon, and B.~Schiele, ``A dataset for movie
  description,'' in \emph{Proceedings of the IEEE Conference on Computer Vision
  and Pattern Recognition}, 2015, pp. 3202--3212.

\bibitem{guadarrama2013youtube2text}
S.~Guadarrama, N.~Krishnamoorthy, G.~Malkarnenkar, S.~Venugopalan, R.~Mooney,
  T.~Darrell, and K.~Saenko, ``Youtube2text: Recognizing and describing
  arbitrary activities using semantic hierarchies and zero-shot recognition,''
  in \emph{Proceedings of the IEEE International Conference on Computer
  Vision}, 2013, pp. 2712--2719.

\bibitem{xu2016msr}
J.~Xu, T.~Mei, T.~Yao, and Y.~Rui, ``Msr-vtt: A large video description dataset
  for bridging video and language,'' in \emph{Proceedings of the IEEE
  Conference on Computer Vision and Pattern Recognition}, 2016, pp. 5288--5296.

\bibitem{rohrbach2017movie}
A.~Rohrbach, A.~Torabi, M.~Rohrbach, N.~Tandon, C.~Pal, H.~Larochelle,
  A.~Courville, and B.~Schiele, ``Movie description,'' \emph{International
  Journal of Computer Vision}, vol. 123, no.~1, pp. 94--120, 2017.

\bibitem{teng1999correcting}
C.-M. Teng, ``Correcting noisy data.'' in \emph{ICML}.\hskip 1em plus 0.5em
  minus 0.4em\relax Citeseer, 1999, pp. 239--248.

\bibitem{kotsiantis2006data}
S.~Kotsiantis, D.~Kanellopoulos, and P.~Pintelas, ``Data preprocessing for
  supervised leaning,'' \emph{International Journal of Computer Science},
  vol.~1, no.~2, pp. 111--117, 2006.

\bibitem{nawi2013effect}
N.~M. Nawi, W.~H. Atomi, and M.~Rehman, ``The effect of data pre-processing on
  optimized training of artificial neural networks,'' \emph{Procedia
  Technology}, vol.~11, pp. 32--39, 2013.

\bibitem{thomason2014integrating}
J.~Thomason, S.~Venugopalan, S.~Guadarrama, K.~Saenko, and R.~J. Mooney,
  ``Integrating language and vision to generate natural language descriptions
  of videos in the wild.'' in \emph{Coling}, vol.~2, no.~5, 2014, p.~9.

\bibitem{cho2015describing}
K.~Cho, A.~Courville, and Y.~Bengio, ``Describing multimedia content using
  attention-based encoder-decoder networks,'' \emph{IEEE Transactions on
  Multimedia}, vol.~17, no.~11, pp. 1875--1886, 2015.

\bibitem{krishnamoorthy2013generating}
N.~Krishnamoorthy, G.~Malkarnenkar, R.~J. Mooney, K.~Saenko, and S.~Guadarrama,
  ``Generating {N}atural-{L}anguage {V}ideo {D}escriptions {U}sing
  {T}ext-{M}ined {K}nowledge.'' in \emph{AAAI}, vol.~1, 2013, p.~2.

\bibitem{zhu2015aligning}
Y.~Zhu, R.~Kiros, R.~Zemel, R.~Salakhutdinov, R.~Urtasun, A.~Torralba, and
  S.~Fidler, ``Aligning {B}ooks and {M}ovies: Towards {S}tory-like {V}isual
  {E}xplanations by {W}atching {M}ovies and {R}eading {B}ooks,'' in
  \emph{Proceedings of the IEEE International Conference on Computer Vision},
  2015, pp. 19--27.

\bibitem{venugopalan2015sequence}
S.~Venugopalan, M.~Rohrbach, J.~Donahue, R.~Mooney, T.~Darrell, and K.~Saenko,
  ``Sequence to {S}equence-{V}ideo to {T}ext,'' in \emph{Proceedings of the
  IEEE International Conference on Computer Vision}, 2015, pp. 4534--4542.

\bibitem{guo2018exploiting}
Y.~Guo, J.~Zhang, and L.~Gao, ``Exploiting long-term temporal dynamics for
  video captioning,'' \emph{World Wide Web}, pp. 1--15, 2018.

\bibitem{xu2018sequential}
Y.~Xu, Y.~Han, R.~Hong, and Q.~Tian, ``Sequential video vlad: training the
  aggregation locally and temporally,'' \emph{IEEE Transactions on Image
  Processing}, vol.~27, no.~10, pp. 4933--4944, 2018.

\bibitem{venugopalan2014translating}
S.~Venugopalan, H.~Xu, J.~Donahue, M.~Rohrbach, R.~Mooney, and K.~Saenko,
  ``Translating {V}ideos to {N}atural {L}anguage using {D}eep {R}ecurrent
  {N}eural {N}etworks,'' \emph{arXiv preprint arXiv:1412.4729}, 2014.

\bibitem{li2018residual}
X.~Li, Z.~Zhou, L.~Chen, and L.~Gao, ``Residual attention-based lstm for video
  captioning,'' \emph{World Wide Web}, pp. 1--16, 2018.

\bibitem{bin2018describing}
Y.~Bin, Y.~Yang, F.~Shen, N.~Xie, H.~T. Shen, and X.~Li, ``Describing video
  with attention-based bidirectional lstm,'' \emph{IEEE Transactions on
  Cybernetics}, 2018.

\bibitem{baraldi2017hierarchical}
L.~Baraldi, C.~Grana, and R.~Cucchiara, ``Hierarchical boundary-aware neural
  encoder for video captioning,'' in \emph{Computer Vision and Pattern
  Recognition (CVPR), 2017 IEEE Conference on}.\hskip 1em plus 0.5em minus
  0.4em\relax IEEE, 2017, pp. 3185--3194.

\bibitem{mikolov2013efficient}
T.~Mikolov, K.~Chen, G.~Corrado, and J.~Dean, ``Efficient estimation of word
  representations in vector space,'' \emph{arXiv preprint arXiv:1301.3781},
  2013.

\bibitem{pennington2014glove}
J.~Pennington, R.~Socher, and C.~Manning, ``Glove: Global vectors for word
  representation,'' in \emph{Proceedings of the 2014 conference on empirical
  methods in natural language processing (EMNLP)}, 2014, pp. 1532--1543.

\bibitem{song2017deterministic}
J.~Song, Y.~Guo, L.~Gao, X.~Li, A.~Hanjalic, and H.~T. Shen, ``From
  deterministic to generative: Multi-modal stochastic rnns for video
  captioning,'' \emph{arXiv preprint arXiv:1708.02478}, 2017.

\bibitem{venugopalan2016improving}
S.~Venugopalan, L.~A. Hendricks, R.~Mooney, and K.~Saenko, ``Improving
  {L}{S}{T}{M}-based {V}ideo {D}escription with {L}inguistic {K}nowledge
  {M}ined from {T}ext,'' \emph{arXiv preprint arXiv:1604.01729}, 2016.

\bibitem{wang2018sequence}
\BIBentryALTinterwordspacing
H.~Wang, C.~Gao, and Y.~Han, ``Sequence in sequence for video captioning,''
  \emph{Pattern Recognition Letters}, 2018. [Online]. Available:
  \url{http://www.sciencedirect.com/science/article/pii/S0167865518303234}
\BIBentrySTDinterwordspacing

\bibitem{li2018multimodal}
W.~Li, D.~Guo, and X.~Fang, ``Multimodal architecture for video captioning with
  memory networks and an attention mechanism,'' \emph{Pattern Recognition
  Letters}, vol. 105, pp. 23--29, 2018.

\bibitem{chen2016video}
T.-H. Chen, K.-H. Zeng, W.-T. Hsu, and M.~Sun, ``Video captioning via sentence
  augmentation and spatio-temporal attention,'' in \emph{Asian Conference on
  Computer Vision}.\hskip 1em plus 0.5em minus 0.4em\relax Springer, 2016, pp.
  269--286.

\bibitem{yang2017catching}
Z.~Yang, Y.~Han, and Z.~Wang, ``Catching the temporal regions-of-interest for
  video captioning,'' in \emph{Proceedings of the 2017 ACM on Multimedia
  Conference}.\hskip 1em plus 0.5em minus 0.4em\relax ACM, 2017, pp. 146--153.

\bibitem{wu2018multi}
A.~Wu and Y.~Han, ``Multi-modal circulant fusion for video-to-language and
  backward.'' in \emph{IJCAI}, 2018, pp. 1029--1035.

\bibitem{yu2018joint}
Y.~Yu, J.~Kim, and G.~Kim, ``A joint sequence fusion model for video question
  answering and retrieval,'' in \emph{The European Conference on Computer
  Vision (ECCV)}, September 2018.

\bibitem{pasunuru2017multi}
R.~Pasunuru and M.~Bansal, ``Multi-task video captioning with video and
  entailment generation,'' \emph{arXiv preprint arXiv:1704.07489}, 2017.

\bibitem{li2018end}
L.~Li and B.~Gong, ``End-to-end video captioning with multitask reinforcement
  learning,'' \emph{arXiv preprint arXiv:1803.07950}, 2018.

\bibitem{pasunuru2017reinforced}
R.~Pasunuru and M.~Bansal, ``Reinforced video captioning with entailment
  rewards,'' \emph{arXiv preprint arXiv:1708.02300}, 2017.

\bibitem{wang2018video}
X.~Wang, W.~Chen, J.~Wu, Y.-F. Wang, and W.~Y. Wang, ``Video captioning via
  hierarchical reinforcement learning,'' in \emph{Proceedings of the IEEE
  Conference on Computer Vision and Pattern Recognition}, 2018, pp. 4213--4222.

\bibitem{wang2018reconstruction}
B.~Wang, L.~Ma, W.~Zhang, and W.~Liu, ``Reconstruction network for video
  captioning,'' in \emph{Proceedings of the IEEE Conference on Computer Vision
  and Pattern Recognition}, 2018, pp. 7622--7631.

\bibitem{krizhevsky2012imagenet}
A.~Krizhevsky, I.~Sutskever, and G.~E. Hinton, ``Image{N}et {C}lassification
  with {D}eep {C}onvolutional {N}eural {N}etworks,'' in \emph{Advances in
  Neural Information Processing Systems}, 2012, pp. 1097--1105.

\bibitem{jackson2018style}
P.~T. Jackson, A.~Atapour-Abarghouei, S.~Bonner, T.~Breckon, and B.~Obara,
  ``Style augmentation: Data augmentation via style randomization,''
  \emph{arXiv preprint arXiv:1809.05375}, 2018.

\bibitem{ghiasi2017exploring}
G.~Ghiasi, H.~Lee, M.~Kudlur, V.~Dumoulin, and J.~Shlens, ``Exploring the
  structure of a real-time, arbitrary neural artistic stylization network,''
  \emph{arXiv preprint arXiv:1705.06830}, 2017.

\bibitem{gatys2016image}
L.~A. Gatys, A.~S. Ecker, and M.~Bethge, ``Image style transfer using
  convolutional neural networks,'' in \emph{Proceedings of the IEEE Conference
  on Computer Vision and Pattern Recognition}, 2016, pp. 2414--2423.

\bibitem{johnson2016perceptual}
J.~Johnson, A.~Alahi, and L.~Fei-Fei, ``Perceptual losses for real-time style
  transfer and super-resolution,'' in \emph{European Conference on Computer
  Vision}.\hskip 1em plus 0.5em minus 0.4em\relax Springer, 2016, pp. 694--711.

\bibitem{vedaldi2016instance}
V.~L. D. U.~A. Vedaldi, ``Instance normalization: The missing ingredient for
  fast stylization,'' \emph{arXiv preprint arXiv:1607.08022}, 2016.

\bibitem{li2018closed}
Y.~Li, M.-Y. Liu, X.~Li, M.-H. Yang, and J.~Kautz, ``A closed-form solution to
  photorealistic image stylization,'' \emph{arXiv preprint arXiv:1802.06474},
  2018.

\bibitem{zhang2015text}
X.~Zhang and Y.~LeCun, ``Text understanding from scratch,'' \emph{arXiv
  preprint arXiv:1502.01710}, 2015.

\bibitem{fellbaum2010wordnet}
C.~Fellbaum, ``Wordnet,'' in \emph{Theory and applications of ontology:
  computer applications}.\hskip 1em plus 0.5em minus 0.4em\relax Springer,
  2010, pp. 231--243.

\bibitem{saito2017improving}
I.~Saito, J.~Suzuki, K.~Nishida, K.~Sadamitsu, S.~Kobashikawa, R.~Masumura,
  Y.~Matsumoto, and J.~Tomita, ``Improving neural text normalization with data
  augmentation at character-and morphological levels,'' in \emph{Proceedings of
  the Eighth International Joint Conference on Natural Language Processing
  (Volume 2: Short Papers)}, vol.~2, 2017, pp. 257--262.

\bibitem{fadaee2017data}
M.~Fadaee, A.~Bisazza, and C.~Monz, ``Data augmentation for low-resource neural
  machine translation,'' \emph{arXiv preprint arXiv:1705.00440}, 2017.

\bibitem{papineni2002bleu}
K.~Papineni, S.~Roukos, T.~Ward, and W.-J. Zhu, ``B{L}{E}{U}: a {M}ethod for
  {A}utomatic {E}valuation of {M}achine {T}ranslation,'' in \emph{Proceedings
  of the 40th annual meeting on Association for Computational
  Linguistics}.\hskip 1em plus 0.5em minus 0.4em\relax Association for
  Computational Linguistics, 2002, pp. 311--318.

\bibitem{lin2004rouge}
C.-Y. Lin, ``R{O}{U}{G}{E}: A {P}ackage for {A}utomatic {E}valuation of
  {S}ummaries,'' in \emph{Text summarization branches out: Proceedings of the
  ACL-04 workshop}, vol.~8.\hskip 1em plus 0.5em minus 0.4em\relax Barcelona,
  Spain, 2004.

\bibitem{banerjee2005meteor}
S.~Banerjee and A.~Lavie, ``M{E}{T}{E}{O}{R}: An {A}utomatic {M}etric for
  {M}{T} {E}valuation with {I}mproved {C}orrelation with {H}uman {J}udgments,''
  in \emph{Proceedings of the ACL Workshop on Intrinsic and Extrinsic
  Evaluation Measures for Machine Translation and/or Summarization}, vol.~29,
  2005, pp. 65--72.

\bibitem{he2016deep}
K.~He, X.~Zhang, S.~Ren, and J.~Sun, ``Deep {R}esidual {L}earning for {I}mage
  {R}ecognition,'' in \emph{Proceedings of the IEEE Conference on Computer
  Vision and Pattern Recognition}, 2016, pp. 770--778.

\bibitem{tran2015learning}
D.~Tran, L.~Bourdev, R.~Fergus, L.~Torresani, and M.~Paluri, ``Learning
  {S}patiotemporal {F}eatures with 3{D} {C}onvolutional {N}etworks,'' in
  \emph{Proceedings of the IEEE International Conference on Computer Vision},
  2015, pp. 4489--4497.

\bibitem{hochreiter1997long}
S.~Hochreiter and J.~Schmidhuber, ``Long short-term memory,'' \emph{Neural
  computation}, vol.~9, no.~8, pp. 1735--1780, 1997.

\bibitem{chung2014empirical}
J.~Chung, C.~Gulcehre, K.~Cho, and Y.~Bengio, ``Empirical {E}valuation of
  {G}ated {R}ecurrent {N}eural {N}etworks on {S}equence {M}odeling,''
  \emph{arXiv preprint arXiv:1412.3555}, 2014.

\bibitem{engstrom2016faststyletransfer}
L.~Engstrom, ``Fast style transfer,''
  \url{https://github.com/lengstrom/fast-style-transfer/}, 2016, commit
  c77c028.

\bibitem{chen2011collecting}
D.~L. Chen and W.~B. Dolan, ``Collecting highly parallel data for paraphrase
  evaluation,'' in \emph{Proceedings of the 49th Annual Meeting of the
  Association for Computational Linguistics (ACL-2011)}, Portland, OR, June
  2011.

\bibitem{xu2015show}
K.~Xu, J.~Ba, R.~Kiros, K.~Cho, A.~Courville, R.~Salakhudinov, R.~Zemel, and
  Y.~Bengio, ``Show, attend and tell: Neural image caption generation with
  visual attention,'' in \emph{International conference on machine learning},
  2015, pp. 2048--2057.

\bibitem{maaten2008visualizing}
L.~v.~d. Maaten and G.~Hinton, ``Visualizing data using t-sne,'' \emph{Journal
  of machine learning research}, vol.~9, no. Nov, pp. 2579--2605, 2008.

\bibitem{kilickaya2016re}
M.~Kilickaya, A.~Erdem, N.~Ikizler-Cinbis, and E.~Erdem, ``Re-evaluating
  automatic metrics for image captioning,'' \emph{arXiv preprint
  arXiv:1612.07600}, 2016.

\end{thebibliography}

\begin{IEEEbiographynophoto}{Silvia Cascianelli} received the M.Sc. \textit{magna cum laude} degree in Information and Automation Engineering in 2015. 
She then joined the Intelligent Systems, Automation and Robotics Laboratory (ISARLab) in 2015 as a Ph.D. student and she is currently a Research Assistant there. Her research interests are mainly Machine Learning, Natural Language Processing, and Computer Vision for Robotics.
\end{IEEEbiographynophoto}
\vspace{-1cm}
\begin{IEEEbiographynophoto}{Gabriele Costante}
received the Ph.D. degree in Robotics from the University of Perugia in 2016. 
He is currently a Post-Doc Researcher at the ISARLab and a Lecturer of Computer Vision at the University of Perugia, Department of Engineering. His research interests are mainly Robotics, Computer Vision and Machine Learning.
\end{IEEEbiographynophoto}
\vspace{-1cm}
\begin{IEEEbiographynophoto}{Alessandro Devo}
received the M.Sc. \textit{magna cum laude} degree in Information and Robotics Engineering in 2018 from University of Perugia, with a thesis on Natural Language Video Description for Service Robotics Applications from the University of Perugia. 
He then joined the ISARlab as a Ph.D. Student. 
His research interests are mainly Machine Learning, Reinforcement Lerning, and Computer Vision.
\end{IEEEbiographynophoto}
\vspace{-1cm}
\begin{IEEEbiographynophoto}{Thomas A. Ciarfuglia} 
received the Ph.D. degree in Robotics from the University of Perugia in 2011. 
He joined the ISARLab in 2008 and worked as a Post-Doc there. He is a Lecturer of Machine Learning at the University of Perugia, Department of Engineering. His research interests are Machine Learning and Computer Vision for Robotics.
\end{IEEEbiographynophoto}
\vspace{-1cm}
\begin{IEEEbiographynophoto}{Paolo Valigi} received the Ph.D. degree from University of Rome “Tor Vergata” in 1991. From 1990 to 1994 he worked with the Fondazione Ugo Bordoni. 
Since 2004 he has been Full Professor at the University of Perugia, Department of Engineering. 
He is currently the head of the ISARLab. 
His research interests are in the field of Robotics and Systems Biology.
\end{IEEEbiographynophoto}
\vspace{-1cm}
\begin{IEEEbiographynophoto}{Mario L. Fravolini}
received the Ph.D. degree in Electronic Engineering from the University of Perugia in 2000. 
He worked as a Research Assistant in the Control Group at the School of Aerospace Engineering, Georgia Institute of Technology, and at the Department of Mechanical and Aerospace Engineering West Virginia University.
Currently, he is an Associate Professor at the Department of Engineering, University of Perugia. 
His research interests include: Fault Diagnosis, Intelligent and Adaptive Control and Biomedical Imaging.
\end{IEEEbiographynophoto}

\vfill

\end{document}